\documentclass[preprint,12pt]{elsarticle}

\usepackage[english]{babel}

\usepackage{algorithm}
\usepackage{algpseudocode}
\usepackage{float}

\usepackage{xcolor}

\usepackage[letterpaper,top=2cm,bottom=2cm,left=2cm,right=2cm,marginparwidth=1.75cm]{geometry}

\usepackage{listings}

\usepackage{amsmath}
\usepackage{amsfonts}

\usepackage{graphicx}
\usepackage{caption}
\usepackage{subcaption}
\usepackage[section]{placeins} % Prevents floats from going beyond sections
\usepackage{booktabs}
\usepackage{tabularx}
\usepackage{tikz}
\usetikzlibrary{shapes.geometric, arrows}
\tikzstyle{startstop} = [rectangle, rounded corners, minimum width=2.5cm, minimum height=0.8cm, text centered, draw=black, fill=red!30]
\tikzstyle{process} = [rectangle, minimum width=2.5cm, minimum height=0.8cm, text centered, draw=black, fill=blue!30]
\tikzstyle{decision} = [diamond, minimum width=3cm, minimum height=0.8cm, text centered, draw=black, fill=green!30, aspect=2]
\tikzstyle{arrow} = [thick,->,>=stealth]
\usetikzlibrary{positioning}

\usepackage{natbib}
\setcitestyle{authoryear,open={(},close={)}}

\usepackage[colorlinks=true, allcolors=blue]{hyperref}

\title{Sophisticated Learning: A novel algorithm for active learning during model-based planning}

\author[1]{Rowan Hodson\corref{cor1}}
\ead{rhodson@laureateinstitute.org}
\author[2]{St John Grimbly\corref{cor1}}
\ead{uct@stjohngrimbly.com}
\author[2]{Evert A. Boonstra}
\author[2,3]{Bruce A. Bassett}
\author[5]{Charel van Hoof}
\author[3]{Benjamin Rosman}
\author[2]{Mark Solms}
\author[1]{Navid Hakimi}
\author[2,4]{Jonathan P. Shock\fnref{fn1}}  % Marking as co-senior
\author[1]{Ryan Smith\fnref{fn1}}  % Marking as co-senior

\address[1]{Laureate Institute for Brain Research, Tulsa, OK, USA}
\address[2]{University of Cape Town, South Africa}
\address[3]{MIND Institute and CSAM, University of the Witwatersrand, South Africa}
\address[4]{INRS, Montreal, Canada}
\address[5]{Delft University of Technology, Department of Cognitive Robotics}

\cortext[cor1]{Corresponding authors.}
\fntext[fn1]{Co-Senior Authors.}

\begin{document}
\begin{abstract} % 250 word limit in Math. Psych. if I recall 
We introduce Sophisticated Learning (SL), a planning-to-learn algorithm that embeds active parameter learning inside the Sophisticated Inference (SI) tree-search framework of Active Inference. Unlike SI -- which optimizes beliefs about hidden states -- SL also updates beliefs about model parameters within each simulated branch, enabling counterfactual reasoning about how future observations would improve subsequent planning.

We compared SL with Bayes-adaptive Reinforcement Learning (BARL) agents as well as with its parent algorithm, SI. Using a biologically inspired seasonal foraging task in which resources shift probabilistically over a 10×10 grid, we designed experiments that forced agents to balance probabilistic reward harvesting against information gathering.

In early trials, where rapid learning is vital, SL agents survive, on average, 8.2 \% longer than SI and 35\% longer than Bayes-adaptive Reinforcement Learning. While both SL and SI showed equal convergence performance, SL reached this convergence 40\% faster than SI. Additionally, SL showed robust out-performance of other algorithms in altered environment configurations.

Our results show that incorporating active learning into multi-step planning materially improves decision making under radical uncertainty, and reinforces the broader utility of Active Inference for modeling biologically relevant behavior.
\end{abstract}

\maketitle

% Include subfiles

\section{Introduction}
\label{sec:intro}

In both biological and artificial systems, decision-making involves a fundamental trade-off: whether to exploit current behavioral strategies or explore the possibility of better strategies. This dilemma is illustrated by animal foraging paradigms, where the choice between exploiting a current food source and exploring for potentially richer alternatives is critically informed by both environmental cues and past experience \citep{charnov1976optimal, stephens1986foraging, webb2025foraging}.  In this context, seeking information to optimize behavioral strategies is an important part of adaptive intelligence, enabling organisms and AI agents alike to reduce uncertainty about their environment. The systematic study of this drive to seek information dates back to early psychological research on curiosity. Berlyne (\citeyear{berlyneCuriosityExploration1966}), for instance, introduced distinctions between different forms of curiosity and established the broader concept as a fundamental motivation for knowledge acquisition. His work demonstrated that organisms display an innate drive to resolve uncertainty and gather information, in some cases independently of primary rewards.

There are now many lines of research on this general topic. For example, an emerging body of work has begun to uncover neural mechanisms associated with exploratory behavior, demonstrating how the brain may assign value to information and guide action selection accordingly \citep{gottliebInformationSeekingCuriosity2013,zajkowski2017causal, chakroun2020dopaminergic,tomov2020dissociable, chou2024influence}. Current work on reinforcement learning (RL) algorithms has also investigated several approaches to encourage information-seeking, from simple heuristics (e.g., initializing unvisited states at high values) to more sophisticated approaches based on upper confidence bounds (UCB), Thompson sampling, and other Bayesian principles \citep{Jaksch2010, houthooft2016vime, bellemare2016unifying, Pathak2017, Russo2018}. Building on this work, intrinsic motivation systems have been shown to successfully guide exploration and learning in both artificial and biological agents, particularly during developmental periods \citep{oudeyerHowEvolutionMay2016}. Additional active learning approaches have been surveyed elsewhere \citep{settles2009active}, each suggesting an agent should optimally be driven to infer and sample from sources of data that would most effectively resolve uncertainty. This is closely related to work on optimal experimental design \citep{mackayInformationBasedObjectiveFunctions1992}, which draws on information-theoretic principles for obtaining maximally informative observations. 

Active Inference (ActInf) is a more recently proposed framework for modeling decision-making under uncertainty. One differentiating feature of this framework is that the drive to resolve uncertainty emerges as an intrinsic feature of its value function, which is in turn derived from a biologically inspired set of first principles. ActInf shares many key features with other prominent frameworks, such as RL, but also differs in important ways. First, it intrinsically assumes partial observability within Markov decision processes, and employs variational inference approaches via variational free energy minimization to approximate Bayes optimality in state inference. Second, its objective function, expected free energy (EFE; denoted as $G$ in the mathematical formalism), is itself derived from variational principles and naturally leads to both reward-seeking behavior and directed exploration (e.g., favoring choices with the greatest outcome uncertainty). Conceptually, EFE quantifies the anticipated `surprise' or uncertainty associated with future states and observations, conditioned on preferences and a particular course of action. Minimizing EFE thus drives agents to select actions that are expected to reduce uncertainty about the world (i.e., yield information), while also moving an agent to states that align with its preferences  (defined more formally in Section~\ref{subsec:formalism}). As mentioned, a key advantage of this formulation is that exploration emerges naturally from the underlying inference process, rather than requiring further additions to a value function. This allows ActInf agents to efficiently navigate uncertain environments by prioritizing actions that both maximize future goal-achievement (aligned with preferences) and minimize uncertainty over states and model parameters.

In recent years, ActInf has been compared against traditional decision models within benchmark machine learning environments \citep{Friston2009, Sajid2021, Fountas2020, Tschantz2020, Millidge2021}. Although its performance in these environments has been context-dependent, on the whole it has been largely comparable to other algorithms. This overlap arises in part because the core motivations behind ActInf - maximizing reward and minimizing uncertainty - are conceptually similar to those found in other agent-based machine learning frameworks. In other words, while the implementation differs, particularly in how it unifies epistemic and instrumental imperatives within a single objective function,  the underlying drives are not unique. Consistent with this, \citet{Sajid2021} demonstrated that ActInf aligns with Bayesian RL when exploration drives are removed. More generally, RL and other agent-based approaches tend to converge toward similar solutions when placed in partially observable environments that benefit from a combination of epistemic drives and model-based planning.  Along these lines, \citet{chou5174669systematic} recently showed that complexity-matched RL and ActInf models explained empirical choice behavior on a 3-armed bandit task with similar accuracy. Yet, Bayesian model selection consistently favored ActInf as the model for which the behavior provided the most evidence.

While promising, certain limitations in current implementations of ActInf have motivated efforts to improve both its performance and scalability. In particular, the computational costs associated with its current variational inference (i.e., message passing) and policy selection approaches would be prohibitive in most real-world applications. This has led to efforts to integrate ActInf with other approaches, including deep learning architectures \citep{Catal2020}, Monte Carlo Tree Search (MCTS) \citep{Fountas2020}, and policy gradient methods \citep{Millidge2021}.

Another limitation is that standard ActInf does not achieve Bellman optimality for policy depths greater than one \citep{DaCosta2023}. To address this, a ``Sophisticated Inference'' (SI) algorithm was more recently developed. This algorithm is Bellman-optimal and solves multi-step planning tasks through a recursive tree search \citep{DaCosta2023}\footnote{SI’s Bellman optimality is limited by the pruning necessary for good performance, as noted by \citet[Figure 1]{Paul_EfficientComputationActiveInference_2023}. Pruning terminates the tree search when the action probability falls below a threshold, but this approximation may cause the agent to miss preferred observations deeper in the tree.}. However, SI has not yet been rigorously compared with other algorithms, and there are clear directions for further development, particularly with respect to active learning drives that have been central to the broader framework.

In this paper, we build on previous work to achieve two main objectives. First, we compare SI to other leading algorithms designed to solve similar problems, including Bayes-adaptive RL (BARL) and a representative upper confidence bound (UCB) heuristic \citep{ucb1}. Second, we introduce and evaluate an extension of SI that incorporates active learning, which we term \textit{Sophisticated Learning} (SL).

To demonstrate the unique planning processes and advantages offered by SL, we compare its performance against the aforementioned algorithms in a novel, biologically inspired environment designed to support multiple strategies for directed exploration.  The results provide novel insights by highlighting the strengths and vulnerabilities of each algorithm. As shown in Section~\ref{sec:results}, SL significantly outperforms all other algorithms tested, and both SL and SI achieve better results than BARL, with or without the addition of UCB. 
 \section{Background}
\label{sec:background}

In this section, we more thoroughly situate our approach within the broader landscape of prior work. We begin by examining the theoretical foundations of ActInf models, focusing on their relationship to established decision-making frameworks. We then explore SI as a key extension to standard ActInf, and discuss its relation to active learning and BARL. This sets the stage for the novel algorithm we propose (SL), which combines insights from each of these previous approaches.

\subsection{Formalism and Notation}
\label{subsec:formalism}

We start by establishing the common formalism underlying both ActInf and BARL. Each framework operates within partially observable Markov decision processes (POMDPs), where an agent must infer hidden states, update its beliefs, and select actions to optimize its objectives. While these approaches differ in several ways, they share a reliance on generative models that represent environmental dynamics.

\paragraph{POMDPs and Generative Model Structure}
In RL, a POMDP provides a formal framework for decision-making under uncertainty, where an agent must infer and reason about hidden states through observations. The framework is traditionally defined as a tuple:
\begin{equation}
(S, O, U, B, A, D, E, R),
\end{equation}
where each element characterizes a specific component of the model.

The state space \(S\) consists of hidden states \(s_t \in S\) that evolve over discrete time steps. These are the true states of the environment. The agent does not observe \(s_t\) directly, but infers it from observations. Observations \(o_t \in O\) are drawn from the observation space \(O\) according to a likelihood model \(A\), such that
\begin{equation}
p(o_t \mid s_t, A) = \text{Cat}(A),
\end{equation}
where \(\text{Cat}(A)\) denotes a categorical distribution parameterized by \(A\), representing the probability of each possible observation given each possible state the agent might find itself in. Actions \(u_t \in U\) are selected from the action space \(U\) and influence state transitions according to a transition model \(B\). Transitions are governed by
\begin{equation}
p(s_t \mid s_{t-1}, u_{t-1}, B) = \text{Cat}(B(s_{t-1} \otimes u_{t-1})),
\end{equation}
where \(\otimes\) denotes the Kronecker product, used to construct the full transition matrix from state-action pairs.

The model also includes a prior distribution \(D\) over initial states, \(p(s_1) = D\), and a prior \(E\) encoding an initial bias over actions at \(t = 1\), which can influence early decision-making. The reward function \(R\) is explicitly defined in conventional RL settings, but this is typically replaced by a preference distribution in the ActInf framework.

In ActInf, this structure is reformulated as a generative model that captures the agent's current beliefs about environmental dynamics. Following \citet{Paul_Isomura_Razi_2024}, the reward function \(R\) is replaced by a preference distribution over observations $p(o|C)$ for a finite time horizon \(T \in \mathbb{N}^+\). Thus, for ActInf, the tuple becomes \((S, O, U, B, A, D, E, T, C)\).

The structure of this generative model leads to the following joint probability:
\begin{equation}
p(o_{1:T}, s_{1:T}, u_{1:T}) = p(A)p(B)p(D)p(E)\prod_{t=1}^T p(o_t \mid s_t, A) \prod_{t=2}^T p(s_t \mid s_{t-1}, u_{t-1}, B),
\end{equation}
where priors over the model parameters and initial state are made explicit, and the agent's prior over actions is encoded as \(p(E)\).

\paragraph{Belief Updating and Parameter Learning}
Given this structure, the agent maintains a belief distribution over states, which is updated recursively as new observations are received. Intuitively, this update combines the likelihood of current observations with predictions based on the previous state estimate. Under a Bayesian framework, this belief update follows:

\begin{equation}
q(s_t) \propto p(o_t | s_t, A) \sum_{s_{t-1}} p(s_t | s_{t-1}, u_{t-1}, B) q(s_{t-1}).
\end{equation}

This approximate posterior distribution over states $q(s_t)$ represents the agent's best estimate of the hidden state given its past experience. Under the mean-field approximation, these posterior beliefs over states follow a more computationally tractable form:

\begin{equation}
q(s_t) = \sigma(\log p(s_t) + \log(o_t \cdot As_t))
\end{equation}
where $\sigma$ represents the softmax function.

Beyond state inference, an agent may also need to learn the transition model \(B\) and/or observation model \(A\), which are typically treated as latent variables. To do so, the agent maintains and updates two types of beliefs: beliefs over hidden states and beliefs over model parameters. Beliefs over states are represented using categorical distributions of the form \(q(s_t) = \text{Cat}(s_t)\), while beliefs about the parameters of the observation and transition models are represented using Dirichlet distributions,
\begin{equation}
q(A) = \text{Dir}(a), \quad q(B) = \text{Dir}(b),
\end{equation}
where the concentration parameters \(a\) and \(b\) encode the agent's beliefs about each of these models, respectively. Learning then occurs through Bayesian updates to these Dirichlet-distributed parameters. As the agent observes state-action-state transitions \((s, a, s')\), it updates its beliefs incrementally:
\begin{equation}
b'_{ijk} = b_{ijk} + \mathbb{I}[s_t = i, a_t = j, s_{t+1} = k],
\end{equation}
where \(b'_{ijk}\) represents the updated count for transitioning from state \(i\) to state \(k\) under action \(j\). An analogous update rule applies to the parameters of the observation model (\(a\)), based on observed outcomes under approximate posteriors over states at each time point.

The formalism presented above establishes how agents can maintain and update structured beliefs about their environment. Bayesian updating in this framework is the foundation for both ActInf and BARL. However, the two frameworks diverge in how these beliefs are used to guide behavior.

ActInf frames decision-making as free energy minimization, selecting actions that minimize expected free energy (EFE). This objective inherently balances goal-directed behavior with information-seeking, unifying exploration and exploitation within a single variational principle.

By contrast, BARL formulates planning as inference in a belief-MDP, where the agent's uncertainty over the environment is treated as part of an augmented state space. Exploration is typically implemented via explicit mechanisms, such as UCB methods, to balance the trade-off between exploration and exploitation.

In the following sections, we examine these approaches in detail. We first explore how ActInf extends variational inference to incorporate future observations and policy selection. We then discuss how BARL constructs and solves belief-space MDPs to handle epistemic uncertainty in environmental dynamics.

\subsection{Active Inference and Expected Free Energy}\label{subsubsec:actinf}
ActInf, sometimes referred to as \emph{standard} or \emph{vanilla} Active Inference in the literature, proposes that agents in an environment with probabilistic state-observation mappings accomplish perception, learning, and action selection through minimization of two related quantities: Variational Free Energy ($\mathcal{F}$) and Expected Free Energy ($G$) \citep{Friston2011, Friston2012}. The Variational Free Energy (VFE) is equivalent to the negative Evidence Lower BOund (ELBO) in variational inference:

\begin{equation}\label{eq:1}
\mathcal{F} = \mathbb{E}_{q(s)}\left[\ln{\frac{q(s|o;\phi)}{p(s|o;\theta)}} - \ln{p(o;\theta)}\right] = -\text{ELBO}
\end{equation}

Here, the value of $\mathcal{F}$ is minimized as the approximate posterior, $q(s|o;\phi)$, approaches the true posterior, $p(s|o;\theta)$, where $\phi$ and $\theta$ parameterize the approximate and true posterior respectively. Thus, the agent can approximate optimal state inference (i.e., perception) by finding the distribution $q$ that minimizes $\mathcal{F}$. Note that $\mathcal{F}$ will also be minimized as the marginal likelihood, $p(o;\theta)$, approaches a value of one. Thus, the agent can also optimize its model (i.e., learning) by finding the parameters $\theta$ that minimize $\mathcal{F}$.

In contrast to perception, action selection in ActInf is guided by minimizing the expected free energy ($G$), an objective that extends variational free energy to future action sequences (policies; $\pi$) that treat observations as random variables \citep{Parr2019, Sajid2021}. While EFE is often presented as a straightforward extension of VFE, recent work suggests that their relationship is more nuanced. Specifically, when considering future states and observations, EFE incorporates terms that capture information-seeking (epistemic) and goal-directed (pragmatic) behavior. A common formulation of EFE is:

\begin{equation}\label{eq:2}
\begin{aligned}
G(\pi) &= \mathbb{E}_{q(o,s|\pi)}\left[\ln{\frac{q(s|\pi)}{p(s|o)}} - \ln{p(o)}\right]\\
       &= \underbrace{-\mathbb{E}_{q(o,s|\pi)}[\ln{p(o|s,\pi)} - \ln{q(o|\pi)}]}_\text{epistemic value}] - \underbrace{\mathbb{E}_{q(o|\pi)}[\ln{p(o)}]}_\text{pragmatic value}.
\end{aligned}
\end{equation}

Intuitively, this formulation balances two key factors: (i) reducing uncertainty about states (epistemic value) and (ii) seeking preferred observations (pragmatic value, encoded in the form of fixed priors over observations; see below). Note that alternative formulations exist, such as the Free Energy of the Expected Future \citep{Millidge2021a}, which differs in its specific implications for information-seeking behavior. This underscores the fact that EFE is not a uniquely defined objective, but rather a family of functionals with varying interpretations and computational properties. Recent work also demonstrates that these formulations are not necessarily equivalent \citep{champion2024reframingexpectedfreeenergy}. We focus our review on the standard formulation of EFE.

It is important to note that the first line of Equation \ref{eq:2} is nearly equivalent to $\mathcal{F}$ in Equation \ref{eq:1}. The difference is that \textit{observations} have been included within the expectation. Thus, $G$ calculates the variational free energy of expected future observations. In a POMDP, these expected observations depend on future states, and transitions between states are dependent on the selected policy. Thus, the agent selects actions that are expected to transition the environment into states that will yield observations that minimize $G$.

The decomposition in Equation \ref{eq:2} makes explicit how EFE drives action selection. For ease of exposition, we will start by unpacking the second term in the second line of Equation \ref{eq:2}, which is often referred to as the \textit{pragmatic} term \citep{smith_tutorial}. As mentioned above, this term drives the agent to seek out the observations that it prefers, or finds most rewarding. This follows from a unique approach to goal-directed choice within ActInf, in which the prior, $\ln{p(o)}$, is used to encode relative preferences (i.e., observations with higher ``probability'' are treated as more rewarding). To make this more explicit, it is sometimes shown as $\ln{p(o|C)}$, where $C$ parameterizes this fixed set of preferences and is clearly distinct from expected observations under a policy, ${p(o|\pi)}$. All else being equal, the agent can thus be thought of as finding a policy that is expected to minimize the difference between its goal (target) distribution and the state-observation pair forecast given its policy. This can be thought of as the agent considering: ``Will this policy take me to states that will most likely generate the observations I want to receive?''

The first term in the second line of Equation \ref{eq:2} , the \textit{epistemic value}, instead quantifies how much an agent expects to learn about states under a given policy. Higher epistemic value corresponds to policies that are expected to lead to greater reductions in uncertainty, naturally leading to exploration. An interesting feature of ActInf is that the term is naturally derived from the free energy formulation. While this resembles directed exploration terms in RL \citep{Mann2012}, it does not need to be added separately from the standard value function. Note also that this is more specifically a form of \textit{state exploration} \citep{Schwart2019}. In other words, it drives agents to reduce uncertainty about states. This is distinct from active learning, which instead drives an agent to update beliefs about model parameters (sometimes called \textit{parameter exploration}; discussed further below). This latter form of exploration is more analogous to that used in standard RL (e.g., taking actions to learn about reward probabilities), primarily due to the fact that RL is more often applied in fully observable environments (i.e., MDPs instead of POMDPs).

To support active learning and parameter exploration (when generative model parameters are not known), the EFE can also be extended to consider beliefs about parameters. For example, when applied to parameters defining the likelihood function, $\theta$, this would yield:

\begin{equation}\label{eq:3}
\begin{aligned}
    G(\pi) &= \mathbb{E}_{q(o,s,\theta|\pi)}\left[\ln{\frac{q(s,\theta|\pi)}{p(s,\theta|o)}} - \ln{p(o)}\right] \\
    &= \underbrace{-\mathbb{E}_{q(o,s|\pi)}\left[\ln{p(o|s,\pi)} - \ln{q(o|\pi)}\right]}_{\text{epistemic value}} - \underbrace{\mathbb{E}_{q(o,s|\pi)}\left[D_{kl}(q(\theta|o,s)||q(\theta))\right]}_{\text{novelty}} \\
    &\quad - \underbrace{\mathbb{E}_{q(o|\pi)}\left[\ln{p(o)}\right]}_{\text{pragmatic value}}.
\end{aligned}
\end{equation}

Here, a new term emerges, often called \textit{novelty}, which measures the change in beliefs about model parameters that would result from expected observations under a policy. High novelty indicates that an observation is expected to significantly refine the agent’s beliefs about how hidden states in its environment generate observations, thereby driving parameter exploration. In practice, this encourages agents to sample from underexplored parts of the environment, making it functionally similar to intrinsic motivation mechanisms in RL that encourage diverse experience sampling.

As touched upon above, beliefs about parameters in discrete settings are typically represented by Dirichlet distributions, allowing the agent to encode uncertainty via concentration parameter counts, $\alpha$. This distribution is given by:

\begin{equation}\label{eq:4}
    p(\theta|\alpha) = Dir(\theta|\alpha) = \frac{1}{\mathcal{B}(\alpha)}\prod_{i=1}^{k}\theta_i^{\alpha_i-1},
\end{equation}

where $\mathcal{B}(\alpha)$ is the multivariate Beta function (which acts as a normalizing constant and is defined using gamma functions). The term $\theta = (\theta_1, \dots, \theta_k)$ is a vector of parameters (e.g., probabilities of outcomes for a categorical distribution) such that $\sum_{i=1}^{k} \theta_i = 1$ and $\theta_i \ge 0$ for all $i$. The term $k$ represents the number of possible discrete outcomes or categories for the variable whose probabilities are given by $\theta$. The vector $\alpha = (\alpha_1, \dots, \alpha_k)$, with each $\alpha_i > 0$, contains the concentration parameters. A higher value of $\alpha_i$ relative to other $\alpha_j$ indicates greater confidence or more evidence accumulated for the parameter $\theta_i$. By updating these concentration parameters with each observation, ActInf agents can dynamically adjust their learning process, balancing the trade-off between consolidating known information about parameters and seeking new experience to increase confidence in their true values.

In summary, the EFE functional drives adaptive behavior by favoring policies that are expected to simultaneously maximize reward (preferred observations) and increase confidence in both states and model parameters. Each of these drives is naturally and dynamically weighted by the magnitude of expected reward and the relative uncertainty about current states and environmental statistics. In practice, these components can also be independently weighted via separate constants to provide additional  flexibility in behavior or to better explain sources of individual variability in studies of humans or other animals \citep{chou5174669systematic}.

While ActInf provides a principled framework for adaptive behavior, practical implementations face significant computational challenges. As touched on above, one key issue is the requirement to evaluate entire pre-specified action sequences (policies) in advance, which becomes infeasible due to the combinatorial explosion of possible policies as the planning horizon increases, and possible decision sequences grow. This is further exacerbated by the high computational cost of variational message passing to update beliefs over states in complex environments, and the reliance on hand-crafted generative models, which can be difficult to specify for real-world tasks. These scalability issues have motivated various extensions to ActInf, including deep learning-based approximations \citep{Catal_Verbelen_Nauta_DeBoom_Dhoedt_2020}, Monte Carlo methods \citep{Fountas_Sajid_Mediano_Friston_2020}, and policy gradient techniques \citep{Millidge_2019}. 
A particularly relevant extension is the SI algorithm mentioned above \citep{Friston2021}, which reformulates the EFE objective using recursive tree search to eliminate the need for exhaustive policy enumeration. SI dynamically refines policies by propagating future information back through a hierarchical planning structure, making it a promising approach for scaling ActInf in real-time decision-making. We now explore SI in more depth.

\subsection{Sophisticated Inference}
\label{subsec:SI}

The SI algorithm extends ActInf to address key scalability challenges in planning by replacing pre-specified sets of possible policies with recursive belief propagation. In other words, unlike standard ActInf, which evaluates all possible action sequences upfront, SI dynamically constructs policies through a tree search process that incrementally propagates and evaluates beliefs about future states and observations.

This recursive approach reframes the EFE minimization problem as a Bellman-like equation \citep{Bellman1958}, explicitly conditioning state inference on actions and observations rather than entire policies. Given an action \(u_t\) at time step \(t\) (omitting model parameter inference for clarity), the recursive formulation of EFE is then:

\begin{equation}\label{eq:5}
\begin{aligned}
    G(u_t, s_t) = \mathbb{E}_{q(o_{t+1}, s_{t+1} \mid u_t, s_t)} 
    &\Big[ \ln p(s_{t+1} \mid o_{t+1}, u_t, s_t) - \ln q(s_{t+1} \mid u_t, s_t) \\
    &\quad - \mathbb{E}_{q(o_{t+1})} \big[ \ln p(o_{t+1}) \big] \\
    &\quad + \mathbb{E}_{q(s_{t+1}, u_{t+1})} \big[ G(u_{t+1}, s_{t+1}) \big] \Big].
\end{aligned}
\end{equation}

This decomposition reveals two critical aspects of SI. The first three terms capture the local epistemic and pragmatic value of an action, quantifying expected information gain and expected reward at the current time step. In contrast, the fourth term recursively propagates future EFE across subsequent time steps, allowing the agent to evaluate the long-term consequences of its actions. Because SI iteratively builds a search tree by expanding high-probability belief trajectories, this can be combined with specific pruning mechanisms that make deep planning computationally feasible while maintaining the primary objective function in standard ActInf. 

\paragraph{Tree Search and Belief Propagation} 
As mentioned above, recursive belief propagation in SI specifically constructs a tree-like structure in belief space, rather than state space, making it particularly well-suited for partially observable Markov decision processes (POMDPs). In more detail, given an initial belief over hidden states, SI considers each possible action \(u_t\) and generates hypothetical future observations \(o_{t+1}\) under the agent’s generative model. Each resulting belief state, \(q(s_{t+1} \mid u_t, o_{t+1})\), is evaluated recursively via Eq.~\ref{eq:5}, forming a branching structure.

To manage computational complexity, SI applies two key pruning mechanisms. First, branches are pruned if the prior probability of transitioning to a future belief state falls below a predefined threshold (e.g., \(p = 0.16\) in the original formulation), ensuring that low-probability trajectories do not consume resources. Second, branches whose EFE is higher than alternatives (i.e. relatively less valuable)  by a predefined threshold are discarded early in the search process, reducing the need to evaluate suboptimal paths in full. By iteratively pruning uninformative or suboptimal action sequences in this way, SI avoids exhaustive policy enumeration while still capturing long-horizon dependencies. This allows the agent to selectively explore policies that are likely to yield high epistemic or pragmatic value. These pruning mechanisms are not unique to SI and have been applied as solutions in both normal ActInf and other algorithms. However, combined with the recursive tree search approach, they offer useful advantages over the original ActInf formulation. At present, this approach remains largely untested against other similar algorithms.

Interpreted psychologically, SI enables an agent to engage in hierarchical counterfactual reasoning about future beliefs and observations. The agent implicitly considers the following sequence:

\begin{quote}
\textit{Given my current beliefs over hidden states, if I took action $u_t$ and transitioned to state $s_{t+1}$, what possible observations $o_{t+1}$ might I receive? How would this update my beliefs about the true hidden state? Given this updated belief, what future observations would I expect if I then took action $u_{t+1}$, and how well would those observations align with my preferences and reduce my uncertainty?}
\end{quote}

This iterative belief update process appears to capture the phenomenology of mental simulation and prospective planning, where decisions are evaluated based on imagined future consequences at different time points in the future.

\begin{figure}[h]
    \centering
    \includegraphics[width=0.7\textwidth]{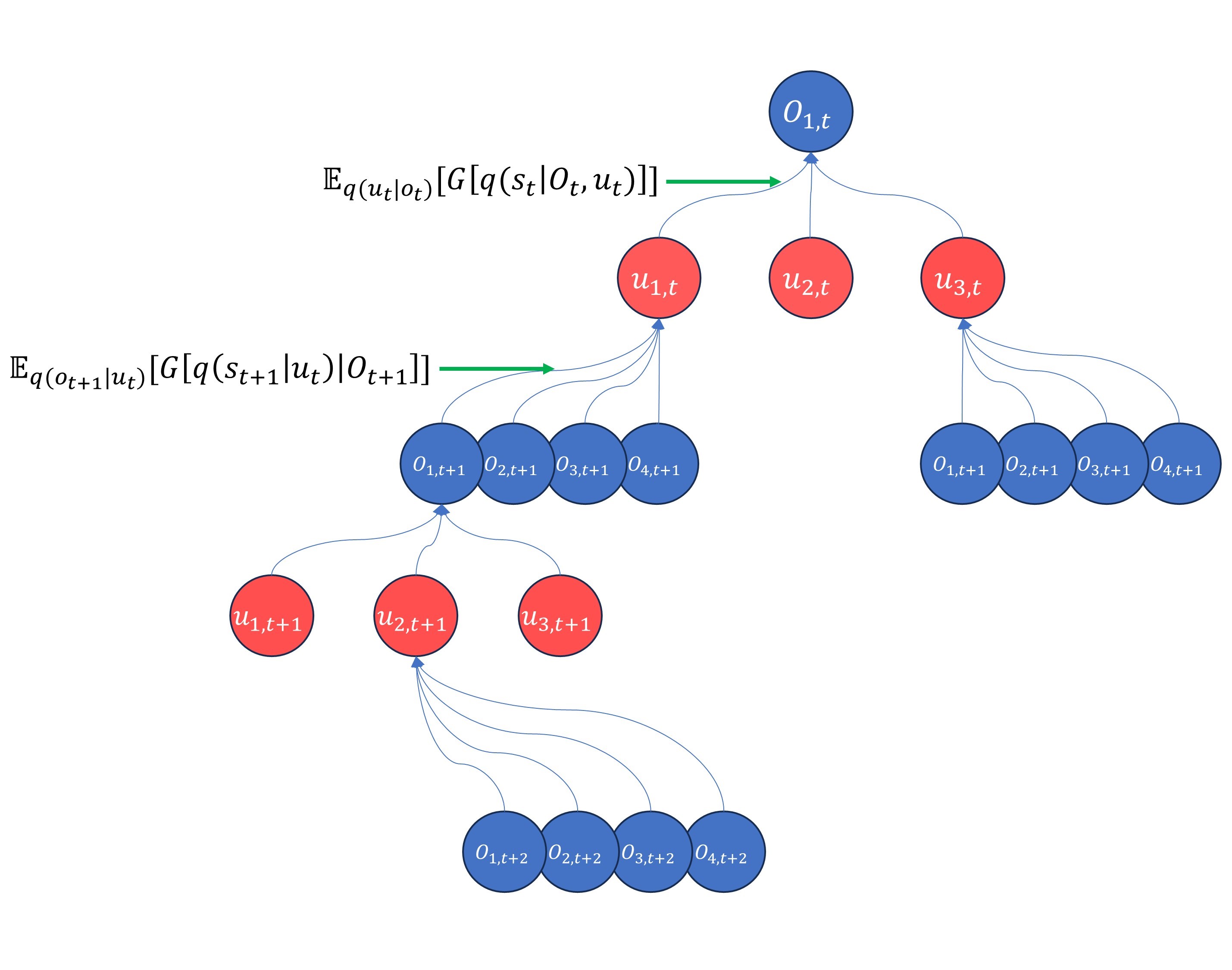}
    \caption{A depiction of the branching recursive search used in Sophisticated Inference (SI). The tree structure shows how beliefs are propagated through sequences of actions ($u$) and observations ($O$), with pruning applied to maintain computational tractability.}
    \label{fig:5by5}
\end{figure}

Figure~\ref{fig:5by5} illustrates this process: each branch corresponds to a candidate action sequence, while belief propagation refines the agent’s expectations about future states and observations.

\subsection{Other Extensions}  
Of note, recent research has also explored various extensions to both standard ActInf and SI. For instance, \citet{Paul_EfficientComputationActiveInference_2023} have proposed the application of dynamic programming techniques to the EFE functional to improve computational efficiency. They also investigate how agents can learn dense preferences over states — representing desirability — by applying Z-learning \citep{Todorov_2006} to a sparse target distribution. This approach enables agents to develop adaptive goal representations beyond predefined reward structures. It also allows the agent to adopt a hierarchical planning perspective, where state preferences emerge dynamically rather than being explicitly assigned. Conceptually, this aligns with intuitive heuristics such as:

\begin{quote}
\textit{“This state will bring me closer to my goal; therefore, I generally prefer this state over the previous one.”}
\end{quote}

However, learning preferences, as opposed to action-based value functions, remains an underexplored area in ActInf, offering new possibilities for adaptive and efficient decision-making.

\subsection{Bayes-Adaptive Reinforcement Learning}
\label{subsec:BARL}

While ActInf frames decision-making through the lens of EFE minimization, an alternative Bayesian approach to handling uncertainty in POMDPs has been described within RL. Specifically, the Bayes-Adaptive Reinforcement Learning (BARL) framework directly extends the classic RL approach by incorporating explicit Bayesian reasoning about model uncertainty, treating the agent's beliefs about environmental dynamics as part of an augmented state space.

Building on the formalism established in Section \ref{subsec:formalism}, BARL also offers a principled approach to exploration by maintaining and updating beliefs over model parameters. Unlike ActInf, which derives information-seeking behavior from EFE minimization, BARL constructs an expanded MDP in belief space, allowing standard optimization techniques to naturally balance exploration and exploitation. This approach has proven particularly effective in scenarios where agents must learn environmental dynamics while maximizing expected rewards \citep{Ross2007}.

\paragraph{Theoretical Foundations}
The BARL approach is situated within the broader field of Bayesian machine learning. To date, considerable work has been carried out in this field, leading to several effective methods for incorporating prior information when performing inference over unknown variables \citep{Ghavamzadeh2015}. These methods are often applied to problems involving uncertainty, where new information is combined with prior beliefs to formulate posterior beliefs about one or more unknown factors. Of particular relevance here, these methods have been effective in navigating POMDPs of the same form assumed by ActInf \citep{Poupart2008}. 

BARL either frames POMDPs with respect to uncertainty over the \textit{solution-space} (model-free), or uncertainty over the \textit{parameter-space} (model-based). A significant advantage of framing such problems in a Bayesian framework is that it effectively side-steps the issue of exploration vs. exploitation. This is due to the fact that Bayesian methods have the capability of representing uncertainty over states/parameters/solutions as \textit{belief} states, which can then be used to identify optimal solutions \citep{Ghavamzadeh2015}. One downside of this approach, however, is its sensitivity to initial priors, which fully determine belief states at the beginning of a task  \citep{Guez2012}. Thus, an integral, and often difficult, aspect of BARL is the design and incorporation of effective prior information.

\paragraph{Model-Based Bayes-Adaptive RL}
Model-based BARL offers a particularly interesting approach to modeling uncertainty over parameters. For example, suppose the true transition probabilities $p(s'|s,a, \theta)$ governing the likelihood of reaching state $s'$ from state $s$ after taking action $a$ are unknown, due to uncertainty in model parameters $\theta$. In this case, the Bayesian agent will maintain \textit{beliefs} about the possible values of $\theta$.

Given $b \in B$, where $B$ is the belief space and $b(\theta) = \text{Pr}(\theta)$, the transition model becomes:

\begin{equation}\label{eq:6}
    p(s'|s,b,a) = \int_{\theta}p(s'|s,a,\theta)b(\theta)d\theta.
\end{equation}

Here, the expectation for $\theta$ with respect to belief $b$ has been taken (i.e., maringalised over), and so $\theta$ does not appear in the resulting probability density. Thus, the model is effectively \textit{known} with respect to belief $b$, and exploration of $\theta$ is not necessary. 

Beliefs themselves are updated upon receiving data (in this case, data about transitions):

\begin{equation}\label{eq:7}
    b' = b(\theta|s,a,s').
\end{equation}
With the model then framed as being known (with respect to $b$), the problem can be formulated as a Markov Decision Process (MDP), and the Bellman equation can be used to determine the optimal value function for each state-belief pair.

\begin{equation}\label{eq:8}
    \upsilon^*(s,b) = argmax_{a}\sum_{s',r}p(s',r|s,a, b)\left[r + \gamma\upsilon(s', b^{s,a,s'})\right]\;\; \forall s \in S,
\end{equation}
where $\gamma$ is a discount factor $\in (0,1]$ which discounts future rewards relative to present ones. As model-based BARL can be formulated as an MDP in this way, an associated algorithm for any normal model-based RL scheme can be constructed. 

Note that, when navigating a fully observable MDP, an agent can learn by registering and storing the number of times it has witnessed specific environmental dynamics. As is implied in equation \ref{eq:7}, this could take the form of an agent increasing its belief that some state $s$ will transition to another state $s'$ after observing the actual occurrence of this specific transition. Its confidence in this transition belief would then increase the more such observations are made. This increase in confidence about a given transition can be represented by incrementing a count ($\mathcal{\phi}^{s'}_{sa}$,) associated with the actual belief over parameters, which is implemented with a Dirichlet distribution using these counts as its concentration parameters. However, when operating within a POMDP, the agent does not fully observe the state-space. Thus, in many cases, it has uncertainty as to what transitions between states actually took place. This creates a scenario where learning is difficult, due to the agent often having inaccurate beliefs about the environmental dynamics underlying its observations. To take this uncertainty into account, the BARL approach to POMDPs incorporates the agent's beliefs over model parameters into the hidden state, forming a \textit{hyper-state} space, $S^* = \big\langle S,T,\mathcal{O} \big\rangle$, with the state-transition and state-observation counts given by $\phi^{s'}_{sa}$ and $\theta^z_{sa}$, respectively. Thus, the space of $T$ and $\mathcal{O}$ is formally defined as:

\begin{gather*}
T = \{\phi \in \mathbb{N}^{|S|^2|A|}, \forall(s,a) \in S \times A, \sum_{s' \in S}\phi^{s'}_{sa}\},\\
\mathcal{O} = \{\theta \in \mathbb{N}^{|S|A|Z|}, \forall(s,a) \in S \times A, \sum_{z \in Z}\theta^{z}_{sa}\}.
\end{gather*}

Given the definitions $T_{\phi}^{sas'} = \frac{\phi^{s'}_{sa}}{\sum_{s\neq s'}\phi^{s'}_{sa}}$ and $\mathcal{O}^{s'az}_\phi = \frac{\theta^{z}_{sa}}{\sum_{z'\neq s'}\phi^{z'}_{sa}}$, the probabilistic dynamics are then:

\begin{equation}\label{eq:9}
    p(s', \phi',\theta',z|s,\phi,\theta,a) = T_{\phi}^{sas'}\mathcal{O}_{\theta}^{s'az}I_{\{\phi'\}}(\mathcal{U}(\phi,s,a,s'))I_{\{\theta'\}}(\mathcal{U}(\theta, s',a,z)),
\end{equation} 
where $\mathcal{U}$ is a function that increases the count of $\phi$ and $\theta$ upon the agent receiving new observations. 

While these equations might appear complex, the concept is simple: Given an initial observation and belief over counts $\phi$ and $\theta$, the agent can, in theory, compute all (countably infinite) hyper-states conditioned on this initial belief. Thus, the model becomes \textit{known} with respect to its priors, with $\mathcal{O}$ and $T$ updated each time the agent receives new data from the environment. It is important to note that, while this represents belief states within a POMDP in a mathematically precise way, convergence is only assured with respect to the agent's initial priors \citep{Katt2018}. However, despite this limitation, the framework has shown good convergence properties in practice \citep{Ross2007, Erik2015, Katt2018}.

\paragraph{Implementation Considerations}
While several choices are possible, the specific BARL algorithm that we consider in simulations below uses online updating, consistent with the approach of \cite{Paquet2005}. Specifically, this version of the algorithm processes data sequentially, updating its beliefs and adapting its strategy incrementally as new information becomes available, rather than requiring the entire dataset upfront. The planning structure (search algorithm) is identical to that used in the SI algorithm, with differences appearing only in the way the reward function is constructed. In general, for these recursive algorithms, search exactly equates to a directed value iteration approach over a subset of reachable states from the initial belief state. 

Algorithmically, the BARL method considered here also simulates searches over the aforementioned hyper-states, which implicitly contain the agent's uncertainty over model parameters. This means the concentration parameter update is performed at every recursive step of the forward tree search (planner), rather than only after every real time-step. For more detailed pseudocode, refer to Algorithm \ref{alg:barl_tree_search_appendix} in the \textit{appendix}. Importantly, the concentration parameter updates during forward tree search are not carried over to the next real time-step - they exist only within the context of the recursive planning. As with SI, the Bayes-adaptive method also implements pruning for both states and actions. 

\paragraph{Comparability of Exploratory Motivations}
As mentioned above, information seeking in BARL follows implicitly from the drive to maximize reward. While this effect is also present in SI, the EFE objective within ActInf contains the \textit{novelty} term as well, which provides a further exploration drive independent of expected reward (i.e., a type of intrinsic curiosity). For greater comparability to SI, BARL can also be supplemented with an explicit directed exploration term. Toward this end, we therefore add a commonly used directed exploration term - an upper confidence bound (UCB) - to BARL in some simulations shown below. Here, UCB takes the form of an algorithmic heuristic that encodes a count over states that an agent has transitioned to up until the current time-point. This can be represented by an expression added to the reward function as follows:

\begin{equation}
    \mathrm{reward} + c\sqrt{\frac{\ln{(t)}}{N_t(a)}}\, .
\end{equation}

% \include{3-1-sophisticated-learning}
% \include{3-2-problem-setting}
% Proposed edited combo of SL and problem setting sections
\section{Methods} \label{sec:methods}

\subsection{Sophisticated Learning}
\label{sec:SL}

% \paragraph{Construction of the Sophisticated Learning (SL) Algorithm}

We now detail construction of the SL algorithm, which integrates SI with insights from active learning and Bayes-adaptive RL. Recall that SI incorporated the first two terms within the EFE decomposition shown in Section~\ref{subsec:SI} within its recursive tree search (i.e., driving state exploration and reward seeking, respectively), but it did not include the third (novelty) term, which serves to motivate parameter exploration. The SL algorithm was specifically constructed to build on SI by incorporating this additional feature, allowing the agent to engage in simulations of potential parameter updates. This specifically allows the agent to reason prospectively about how different future actions are expected to refine its model parameters - thereby improving its ability to learn in dynamic environments.

SL unifies SI and Bayes-adaptive methods, leveraging their respective strengths. As shown below (Section~\ref{sec:results}), both SI and BARL show comparatively poor performance in scenarios requiring complex, adaptive learning. While SI has not been widely tested in such environments \citep{Friston2021}, it is well documented that BARL approaches to POMDPs are highly dependent on well-specified prior beliefs to facilitate effective learning \citep{Ross2007, Katt2018}. This limitation often constrains their applicability in highly uncertain, non-stationary settings, consistent with the results presented here.

By propagating parameter updates within the recursive tree search itself, similar to Bayes-adaptive methods, SL enables an agent to anticipate how its beliefs will evolve over time, rather than treating them as static. This allows the agent to select actions not only for immediate goal optimization but also to maximize its future learning potential. In effect, SL empowers agents to perform counterfactual reasoning about their own epistemic progress, thereby making proactively self-improving decisions that accelerate model convergence and adaptability.

In more detail, the SL algorithm updates concentration parameter counts after every simulated time step, in the same manner as in BARL. These updated concentration parameters are then propagated forward and used to construct (via normalization) the transition and/or likelihood functions, which are used in subsequent steps of recursive search. The SL algorithm can therefore consider how model parameters \textit{would} change along its forward tree search \textit{if} it were to take one action sequence versus another. This is important, as it more adequately represents a simulation of the way an actual real-time trajectory would unfold if the agent were to take a particular set of actions and, in doing so, update its model parameters after every real time step. Note that simulating how states and model parameters change in this way is necessarily based on the agent's prior beliefs over states and model parameters, which can result in incorrect and biased assumptions about the environment. However, such techniques have nonetheless shown good convergence properties \citep{Ross2007}. 

In addition to this method of counterfactual search, SL also implements a ``backward-smoothing'' function - a feature previously suggested (in a more limited scope) in the original presentation of SI \citep{Friston2021}. This backward-smoothing function backtracks from the current evaluated time step to adjust its posterior beliefs over states at previous time steps. This is particularly useful in the case of learning, as it allows for retrospective assignment of observations to posteriors over states, which can result in more accurate updates to the associated Dirichlet concentration parameter counts. Importantly, this backward smoothing function is implemented at each evaluated future time step within the agent's planning horizon, as well as at each real time step.  

In summary, there are two key differences between SL and the original SI scheme. The first is the addition of propagating parameter learning through forward-looking simulations. The second is simulated backward smoothing of parameter learning at every step in this forward search. Psychologically, one could therefore consider an SL agent reasoning as follows: 

\begin{quote}
\textit{If I were to take an action, receive an observation, and transition to a new state, how would I then update my posterior beliefs over states for this time-step and for previous time-steps? Based on these posterior updates, how would I then change my current model?}
\end{quote}

This method of multi-level counterfactual thinking proves particularly beneficial when the agent needs to learn the likelihood function while the state-transition function is known, as described in our primary algorithm comparisons below (Section \ref{primary-methods}). 

While the principle of backward smoothing to refine posteriors over past states exists in other inference schemes, the distinct advantage of SL lies in its proactive integration of this process within its forward planning. Specifically, the search mechanism within SL evaluates and prioritizes actions not only on immediate outcomes but also on the anticipated information gain about parameters that would be achieved through subsequent backward smoothing. It therefore more highly values trajectories leading to states from which backward inference will yield more precise and informative updates to past beliefs, and consequently to the model parameters themselves. As we demonstrate below, this strategic emphasis on future epistemic refinement via backward smoothing contributes to more accurately revised historical beliefs, which in turn supports robust future decision-making and accelerates learning in uncertain environments.
 
\begin{figure}[h]
\centering
\scalebox{0.8}{
\begin{tikzpicture}[node distance=1.5cm and 1.5cm]

% Nodes with short labels
\node (start) [startstop] {Initialise beliefs and parameters};
\node (outerloop) [process, below of=start] {Iterate time steps ($t+1$ to $\tau$)};
\node (updatep) [process, below of=outerloop] {Compute cumulative transition probabilities};
\node (innerloop) [process, below of=updatep] {Iterate over all states};
\node (updatestate) [process, below of=innerloop] {Update beliefs using simulated observations};
\node (endloop) [decision, below of=updatestate, yshift=-1cm] {More timesteps?};
\node (normalise) [process, right of=updatestate, xshift=7cm] {Normalise updated belief distribution};
\node (stop) [startstop, right of=endloop, xshift=7cm] {Return smoothed beliefs};

% Arrows
\draw [arrow] (start) -- (outerloop);
\draw [arrow] (outerloop) -- (updatep);
\draw [arrow] (updatep.south) -- (innerloop.north);
\draw [arrow] (innerloop.south) -- (updatestate.north);
\draw [arrow] (updatestate) -- (endloop);
% \draw [arrow] (endloop.east) -- ++(2,0) |- (normalise.west);

% Decision arrows with labels
\draw [arrow] (endloop.east) -- ++(2,0) |- (normalise.west)
    node[pos=0.25, right]{\textbf{No}};
\draw [arrow] (endloop.west) -- ++(-3,0) |- (outerloop.west)
    node[pos=0.25, left]{\textbf{Yes}};

\draw [arrow] (normalise.south) -- (stop.north);
\draw [arrow] (endloop.west) -- ++(0,0) -- ++(-3,0) |- (outerloop.west);

\end{tikzpicture}
}
% \caption{Flowchart for the Backwards Smoothing Algorithm. 
% The process starts with initialising the state probabilities (`Initialise`). 
% The algorithm then iterates over future time steps (`Iterate timesteps`) while adjusting probabilities based on actions (`Update with actions`). For each timestep, all possible states are checked (`Check states`), and state probabilities are updated using observations (`Update with observations`). 
% If there are more timesteps to process (`All steps done?`), the loop continues. 
% At the end, the probabilities are normalised (`Normalise`) and the final result is returned (`Return result`).}
\caption{Flowchart representation of the Backward Smoothing Algorithm. The algorithm begins by initializing the posterior belief distribution \( L \) and a transition probability accumulator \( p \). It iterates over future time steps \( t+1 \) to \( \tau \), updating \( p \) using the transition function \( p \gets b(\textit{action\_history}(time step - 1)) \times p \). For each state, beliefs are updated according to the likelihood of the observations and transitions as \( L(state) \gets L(state) \times \textit{observation}(time step) \times a \times p(state) \). If there are more time steps to process, the loop continues. Otherwise, the updated belief distribution is normalized and returned as the final smoothed beliefs. Pseudocode is available in the appendix.}
\label{fig:backwards_smoothing}
\end{figure}

\subsection{The Foraging Grid-World Environment}\label{subsec:envir}

To evaluate the relative performance of SL, SI, and BARL, we designed a challenging grid-world environment to test multi-step planning where strategic exploration is essential for maximizing long-term reward. While other environments have been used to compare ActInf to different machine learning algorithms \citep{Sajid2021, Millidge2021}, they often isolate specific behaviors like exploration or model learning. Our environment integrates these demands, requiring agents to dynamically balance exploration, parameter learning, and reward optimization while forecasting probabilistic changes in the world. This design was motivated by common biological challenges: managing distinct and growing needs (e.g., hunger, thirst), avoiding critical survival thresholds, and locating resources whose availability changes over time, necessitating epistemic foraging.

\subsubsection{Environment Details and Agent Model}
\begin{figure}[h]
\centering
\includegraphics[width=0.7\textwidth]{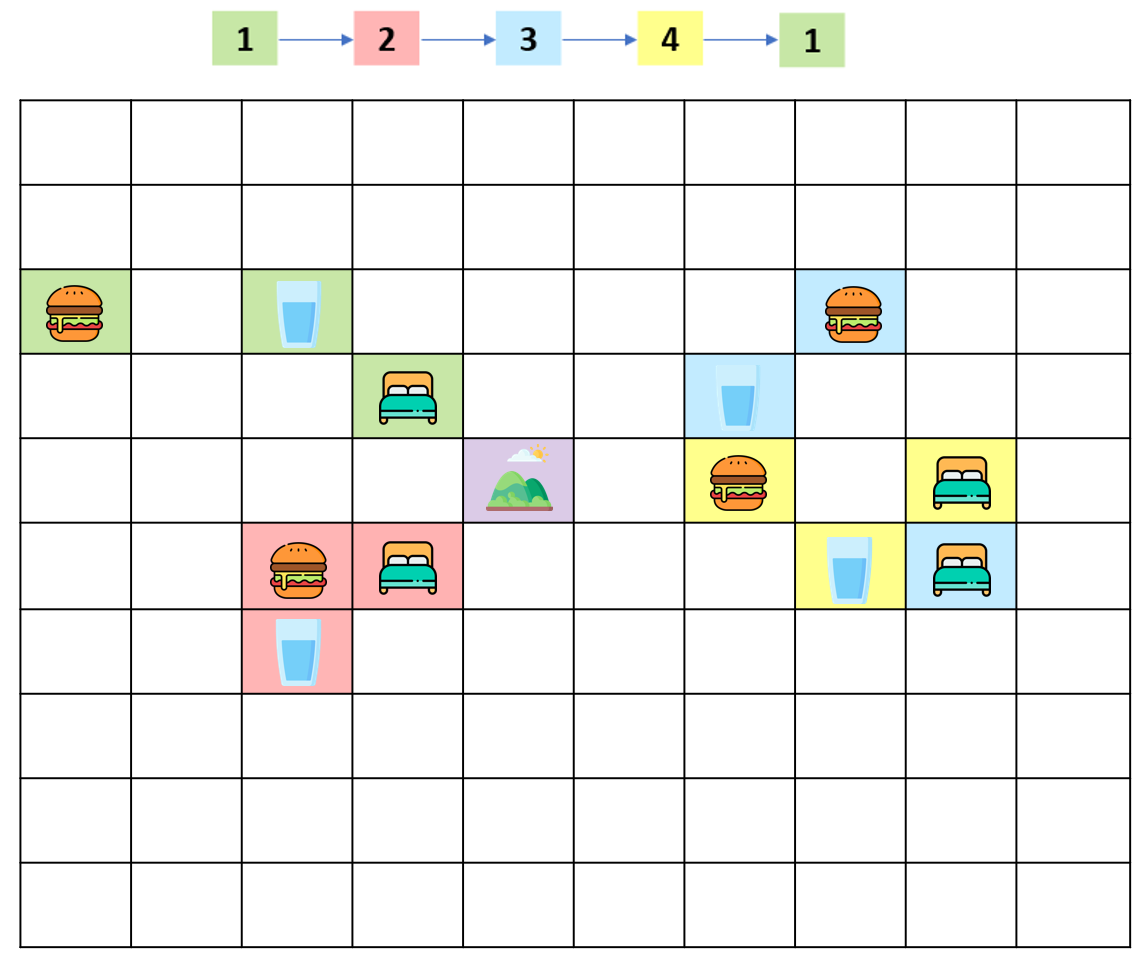}
\caption{An example 10-by-10 grid world in one of four different contexts (with different colors representing different seasonal contexts) defining resource locations.}
\label{fig:gridworld-visualisation}
\end{figure}

The environment is a 10-by-10 grid containing three non-depleting resources which are nominally labeled \texttt{food}, \texttt{water}, and \texttt{sleep} (see Figure~\ref{fig:gridworld-visualisation}). At each time step, an agent can move up, down, left, right, or remain in place. Positional transitions are deterministic and known to the agent.

The core challenge posed by this environment lies in its partially observable nature. The locations of the resources depend on hidden \textbf{context} states which change probabilistically over time. For conceptual purposes, we labeled these context states as seasons (i.e., \texttt{Spring}, \texttt{Summer}, \texttt{Autumn}, \texttt{Winter}). The agent could not directly observe season states. However, it could temporarily reveal the current season by visiting a specific cue location we refer to as the \textbf{Hill} state (i.e., as if providing an overview of the environment). However, visiting the hill state did not reveal the resource locations themselves. Thus, the agent was still required to learn the mapping between seasons and resource locations through exploration. This setup created an explicit explore-exploit dilemma in which the agent was required to choose between: 1) exploring new locations to find resources, 2) visiting the hill to reduce uncertainty about the current season, or 3) exploiting current beliefs and moving toward locations with previously observed resources.

The agent's state space was formally defined as:
$$ \text{position} \times \text{food}_t \times \text{water}_t \times \text{sleep}_t \times \text{context} $$
Here, $\text{food}_t$, $\text{water}_t$, and $\text{sleep}_t$ are internal states tracking the time steps since each resource was last acquired. These functioned as homeostatic needs that grew over time, where each resource level was known to the agent with certainty\footnote{One might consider making internal need states uncertain as well, which we note as an interesting possible direction for future work.}.

Formally, the agent had two observation modalities. The first pertained to resources within a grid state, with four possible outcomes: \texttt{Empty}, \texttt{Food}, \texttt{Water}, or \texttt{Sleep}. The second modality provided information about the context. Namely, if at the hill state, the agent observed the current context (e.g., \texttt{Winter}), while all other grid locations provided a non-informative \texttt{No Context} observation.

\subsubsection{Dynamic Multi-Objective Preferences}
As mentioned above, preferences in this environment were not static; they were determined by a dynamic, multi-objective reward function that reflected the agent's current internal resource needs (Algorithm~\ref{alg:cap}). The preference for a given resource increased as the time since its last acquisition grew. If any resource timer exceeded a predefined limit, the agent incurred a large penalty and the trial ended. In some ways, this outcome could be thought of as an agent's `death' (although, as described below, learning was allowed to carry over between trials for evaluative purposes). This structure, inspired by homeostatic regulation, forced the agent to balance multiple competing objectives to ensure survival, a design that follows classical approaches in reinforcement learning \citep{Sutton2018}.

\begin{algorithm}
\caption{Multi-objective reward function}\label{alg:cap}
\begin{algorithmic}
\Function{Preferences}{$resources_t$, $resources_L$, penalty}
    \State $empty_p$ $\gets$ -1
    \For{each resource in [water, food, shelter]}
        \State $resources_p$ $\gets$ $resource_t$ 
        \If{$resource_p$ $\geq$ $resources_L$}
            \State $empty_p$ $\gets$ penalty
            \State $resources_p$ $\neq$ $resource_L$ $\gets$ penalty
        \EndIf
    \EndFor\\
    \Return [$empty_p, resources_p$]
\EndFunction
\end{algorithmic}
\end{algorithm}
\begin{figure}[h]
\centering
\scalebox{0.8}{
\begin{tikzpicture}[node distance=1.5cm and 2cm]

% Nodes
\node (start) [startstop] {Start: Set penalties};
\node (resource_loop) [process, below of=start] {Iterate over resources};
\node (check) [decision, below of=resource_loop, yshift=-0.8cm] {Is resource sufficient?};
\node (apply_penalty) [process, right of=check, xshift=5cm] {Apply penalty};
\node (next_resource) [process, below of=check, yshift=-1cm] {Next resource};
\node (return) [startstop, below of=next_resource, yshift=-1cm] {Return values};

% Arrows
\draw [arrow] (start) -- (resource_loop);
\draw [arrow] (resource_loop) -- (check);
\draw [arrow] (check.east) -- ++(0.5,0) |- (apply_penalty.west);
\draw [arrow] (apply_penalty) |- (next_resource.east);
\draw [arrow] (check.south) -- (next_resource.north);
\draw [arrow] (next_resource) -- (return);

\end{tikzpicture}
}
\caption{Flowchart for the multi-objective reward function, which sets an agent's preferences at time $t$. The process starts by setting penalties. 
It then iterates over all resources (water, food, sleep). For each resource, the algorithm checks if it meets the required threshold. 
If it does not, a penalty is applied. The process continues to the next resource, and after processing all resources, it returns the final values.}
\label{fig:multi_obj_reward}
\end{figure}

The agent's dynamic preference structure was a key feature of this task. Unlike typical ActInf implementations with static or solely time-dependent preferences \citep{Tschantz2020, Sajid2021, Friston2021, smith_tutorial}, here the agent's preferences were a function of its own policy. Namely, the actions an agent took determined its future internal states, which in turn defined its future preferences. This created a circular dependency in which the agent was required to identify a policy that best satisfied the very preferences induced by that policy.

\subsubsection{Illustrative Task Example}

The  design of this environment enabled nuanced, non-trivial strategies for uncertainty reduction. While much of the extant work on epistemic behavior has focused on bandit tasks \citep{averbeck2015theory, MarkovićBandits}, our environment instead allowed for long-term sequential planning. For example, an agent could infer the current context in two distinct ways: directly, by visiting the hill, or indirectly, by visiting a location where a resource was known to exist in a specific context. Observing the resource confirmed the context, while its absence implied the context had changed.

\begin{figure}[h]
    \centering
    \includegraphics[width=\linewidth]{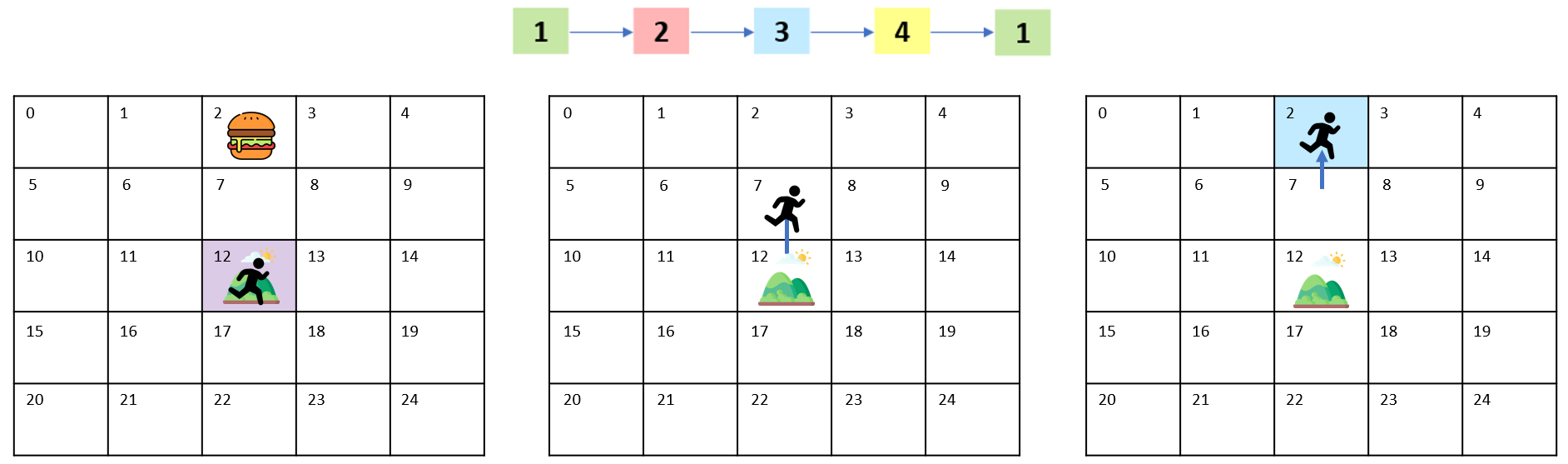}
    \caption{Illustration of an agent’s reasoning scenario within a \(5 \times 5\) grid-world. (A) The agent starts at the hill (position 12) at \( t = 0 \). (B) The agent moves to position 7 at \( t = 1 \). (C) The agent then reaches position 2 at \( t = 2 \). At this position, it expects to find food if the context remains \texttt{Winter}. The absence of food allows the agent to infer that a probabilistic transition has occurred between seasons, guided by its model of cyclic seasonal change.}
    \label{fig:bayesian_inference_example}
\end{figure}

To illustrate, consider the scenario in Figure~\ref{fig:bayesian_inference_example}. In this example, the transition probabilities between seasons were known, but the mappings between grid locations and resources in each season (i.e., the likelihood function) needed to be learned.  At $t=0$, this simulation assumed the agent was at the hill state and observed the context was \texttt{Winter}. It also assumed the agent had previously learned through experience that \texttt{Food} was likely at grid position 2 in \texttt{Winter}. The agent therefore moves for two time steps to reach position 2. 

If the probability that the season would remain \texttt{Winter} was $0.95$ per time step, the probability of it still being \texttt{Winter} upon arrival would be $0.95 \times 0.95 = 0.9025$. Thus, the agent remains fairly confident the season has remained stable. However, when the agent arrives at position 2, it finds that food is absent. This allows the agent to confidently infer the season had changed. As the agent knows the transition probabilities between seasons, it could also reason about the most likely context transition when updating its beliefs (e.g., a single transition to \texttt{Spring} vs. a double transition to \texttt{Summer}). This example highlights how optimal behavior requires an agent to rely on its model of the world to guide belief updating and guide action selection toward exploratory or reward-seeking choices.

\subsection{Experiment Setup and Details}\label{subsec:setup}
For our primary simulations below, the hill state was kept at position 55 (center of the grid) to ensure it was generally within the planning horizon (search-depth) of the agent from all points in the grid. Resource locations were also chosen heuristically, fixed within each season to ensure points of interest were mostly within a reasonable search depth of each other and could effectively contribute to learning. Specifically, depending on the season, \texttt{Food}, \texttt{Water}, and \texttt{Sleep}, were respectively placed in the following positions: \texttt{Spring} = 71, 73, 64; \texttt{Summer} = 43, 33, 44, \texttt{Autumn} = 57, 48, 49, and \texttt{Winter} = 78, 67, 59. Seasonal context transitions remained stable with a 0.95 probability, or transitioned to the adjacent context with a 0.05 probability. The initial context was uniformly sampled at the start of each trial with the agent having a uniform belief over contexts at the first time step. Note that, while we focus on the specific configuration described here for detailed illustration, each algorithm was also tested on several other configurations (i.e., choice of resource locations) to ensure the generalizability of our results. The results of these further confirmatory analyses are provided in \textit{appendix} Section \ref{subsubsec:grid-experiments}. 

Each trial began with the agent at a fixed initial position (state 51). Survival thresholds for food, water, and sleep were set at 22, 20, and 25 time steps, respectively, below which the agent would die (ending the trial). These time-step limits were chosen heuristically to allow the agent sufficient time to learn the model while also mimicking the fact that different resources deplete at different rates in true biological organisms. These limits also prevented selection of behaviors that, while intelligent, were problematic for the questions we aimed to answer (e.g., if too long, the agent would  simply wait in one location the entire time until the season would return where a known resource was present). The preference structure assigned values to observations based on these resource timers, scoring empty states at -1 and resource states positively according to elapsed depletion times ($food_t$, $water_t$, $sleep_t$). Preferences uniformly shifted to a large negative penalty (-500) for all observations upon exceeding any resource threshold.

We defined a \textit{trial} as a single `run-through' of the agent in the environment. Each trial was terminated either upon resource depletion (agent death) or at a specified maximum number of time steps (100 in the case of our experiments). Several trials were conducted in sequence, where any learning in a given trial was carried over to the start of the next trial. We refer to each of these sequences of trials as an \textit{evaluation}. Unlike commonly presented implementations in which parameter values are updated after each full trial \citep{Friston2021}, our implementation performed these updates after every time step. This was necessary for the agent to solve the problems posed by this environment. Thus, all algorithms here operate in a dynamic, `online' manner. %All simulations and analyses were implemented in MATLAB (version R2022b), executed on the University of Cape Town’s high-performance computing cluster. We now present the two main experiment categories which were conducted.

\subsubsection{Analysis of Search Heuristic and Horizon Depth}
As a baseline characterization of performance, we first analyzed both SI and BARL in a setting where all elements of the environment were known—that is, where the model was equipped with full knowledge of transition probabilities between seasons, resource positions, and the location of each resource in each season. 

Note that because model parameters are fixed in this setting and no learning is required, SL collapses to SI and would not offer additional insights if tested in this setting. BARL also collapses to standard Bayesian RL.

For further insights, simulations within a known environment were also carried out under different planning horizons (from 1-9 steps) and with three different tree-search heuristics. This allowed us to identify theoretically optimal depth and search strategies when model learning was not required.  Specifically, we evaluated SI under (i) depth-limited recursive search with memoization, (ii) Monte Carlo roll-outs (with random action selection) and (iii) a hybrid scheme that applied recursive tree-search with memoization to the first \textit{h} steps and Monte Carlo rollouts to the remaining \textit{m} steps (with \textit{h} + \textit{m} = 6). While memoization accelerates inference by caching estimates for previously visited state configurations, it can also sometimes introduce inaccurate cached values. Monte Carlo approaches help to avoid this bias by drawing independent roll-outs (100 in these evaluations) from each leaf node at the expense of greater computational cost. The hybrid approach trades off these properties by reusing exact sub-trees early in the search while relying on unbiased roll-outs deeper in the horizon. Note that this hybrid approach is similar to a partially observable monte-carlo planning approach \citep{silver2010monte}. 

\subsubsection{Primary Algorithm Comparisons}
\label{primary-methods}
After completing the above-mentioned baseline performance characterizations, our primary analyses compared SL to SI and BARL (both with and without UCB). Here, we focused on the case where the likelihood (i.e., the resource positions within each context) needed to be learned and the transition probabilities between seasons were known. The number of time steps an agent survived in each trial, and how this changed across trials in each evaluation, were taken as our primary performance measures. Performance comparisons were initially conducted with 200 sequential trials per evaluation. This length was selected as a computationally reasonable upper bound that allowed for sufficient exploration. These simulations were carried out using a fixed horizon of 9 and a full-depth tree search with memoization. This choice was motivated in part by initial results of the analyses described in the previous section (for results, see \ref{horiz_analysis_res}), which indicated that performance continued to improve up to this horizon. We were also primarily interested in comparison between algorithms when minimally constrained by limitations of  choice of search strategy.

To provide a generalizable characterization of performance, we carried out 500 evaluations of these trial sequences (with 500 random seeds). Following convergence analyses, which indicated that average performance results typically stabilized around 100 trials, evaluations were reduced to 120 trials. This shorter trial number was chosen as it still captured the core learning dynamics post-convergence while significantly reducing computational demands of simulation. This permitted an increase to 2000 seeds for these more extensive analyses to ensure greater statistical confidence and thorough exploration of the patterns of behavior shown by each algorithm. 

To better quantify algorithm performance, we fit Linear Mixed-Effects models (LMEs) using trial, algorithm, and their interaction as predictors of survival time: 

$Survival \sim Trial + Algorithm + Trial \times Algorithm + (1 | Id)$

To evaluate early learning dynamics, we ran these LMEs separately for two key trial intervals: a \emph{Ramp-up Phase} (Trials 1–20) and an \emph{Active Learning Phase} (Trials 21–60).  These models allowed us to estimate both learning rates (slopes) and performance levels (based on estimated marginal means [EMMs]). As an auxiliary feature, to better understand how learning was shaped during experiments, we measured the extent to which SL's model deviated or adhered to the true environment, via KL-divergence analyses.

Additional experiments across varied grid configurations (\textit{appendix} Section \ref{subsubsec:grid-experiments}) were subsequently carried out to more thoroughly compare each of the four algorithms (SL, SI, and BARL with and without UCB). These evaluations were conducted over 200 trials each to maintain consistency with the initial longer runs in our main simulations and provide a comparable basis for assessing performance across different algorithmic approaches. For these multi-algorithm comparisons, 200 seeds per condition were utilized, chosen as a practical balance between computational resources and the need for reliable comparative data across the varied configurations (i.e., varied resource locations by season).
 % Proposed edited combo of SL and problem setting sections
\section{Results} \label{sec:results}

Below we present the results of our two main experiments, along with analysis of major behavioral patterns and underlying mechanisms.

\subsection{Analysis of Search Heuristic and Horizon Depth} \label{horiz_analysis_res}
Figure \ref{fig:memory-tree-search-analysis} displays the results of simulations across different planning horizons and search heuristics when both the likelihood and transition probabilities were known (i.e., serving as an assessment of each algorithm's maximum performance level). Findings indicated that the non-memoization approach outperformed the memoization approach at a horizon of 5 and above, as it avoided the use of potentially inaccurate cached values, albeit at a substantially higher computational cost (approximately 28 times greater than that of the memoization condition at horizon 5). The hybrid search method showed better performance at shorter horizons. However, its relatively inefficient sample usage rendered it less computationally feasible. 

One interesting observation was that BARL showed better performance than SI, with the most notable differences occurring in earlier trials. This was most likely driven by SI's use of the epistemic value term within EFE, which encouraged moving to the hill more frequently. While this can be beneficial during learning, it may detract from reward maximization behavior when the environment is fully known (as in these simulations). 

\begin{figure}[htbp]
  \centering
  % reference width = \linewidth, alignment = centred
  \makebox[\linewidth][c]{%
    \includegraphics[width=1.1\linewidth]{figures/figures_stjohn/figure_6_SL.png}%
  }
  \caption{Analyses testing effects of planning horizon and possible tree-search strategies. \textbf{A} and \textbf{B} show the average number of time steps survived and posterior calculations, respectively, for the Sophisticated Inference algorithm when both likelihood and transition probabilities were known. \textbf{C} and \textbf{D} show similar plots for the Bayesian RL algorithm. Here, `Memory' indicates the use of memoization caching of previously calculated values, `No Memory' indicates the absence of any memoization, `Monte Carlo' indicates the use of Monte Carlo (MC) sampling instead of recursive search, and `Hybrid' indicates the combined use of recursive search and MC sampling. For the hybrid algorithm, combinations of recursive search + MC roll-outs were tested, each with a combined horizon of 6. In other words, this hybrid plot shows the search depth with memoization from 1 (search depth of 1, MC roll-out of 5) to 6 (search depth of 6, MC roll-out of 0).}
  \label{fig:memory-tree-search-analysis}

\end{figure}
\subsection{Relative Performance under Model Uncertainty}
\label{sec:relative_performance_phased}

\begin{figure}[h]
  \centering
  \includegraphics[width=0.7\textwidth]{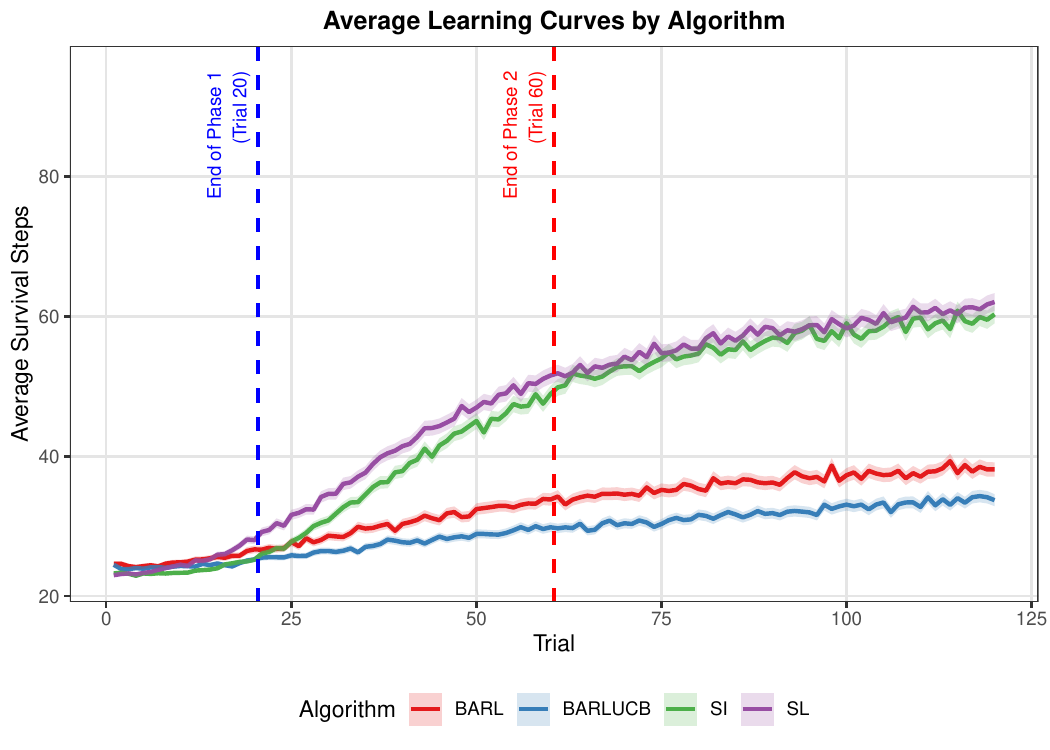}
  \caption{Mean survival per trial across 120 trials by algorithm (shaded bands show 95\% CIs, aggregated over up to 2000 seeds). Vertical dashed lines delineate Phase 1 (Trials 1–20), Phase 2 (Trials 21–60), and Phase 3 (Trials 61–120).}
  \label{fig:explorative}

  \vspace{0.5cm}
  \begin{minipage}{0.49\textwidth}
    \centering
    \includegraphics[width=\linewidth]{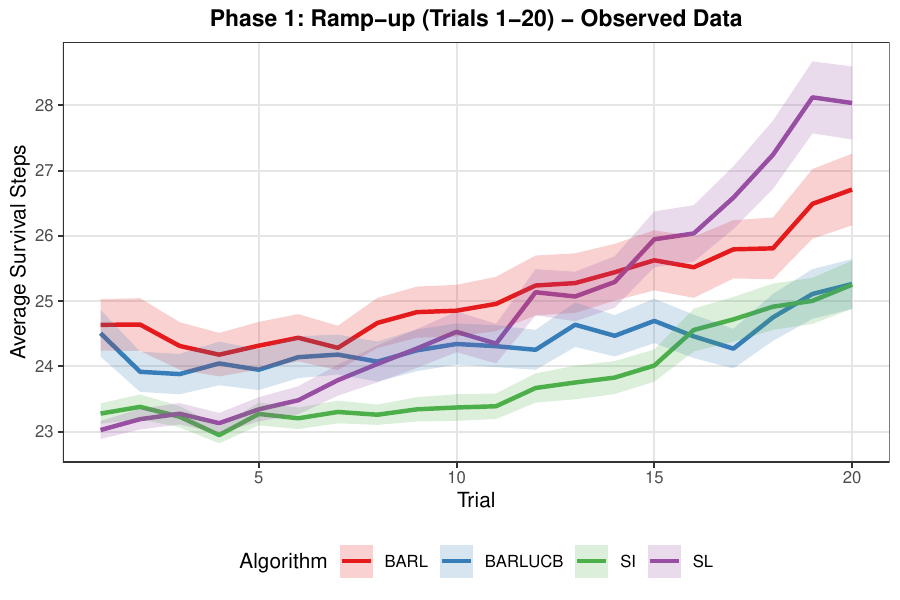}
    \captionof{figure}{Phase 1: Ramp-up (Trials 1–20). Mean observed survival with 95\% CIs.}
    \label{fig:phase1}
  \end{minipage}
  \hfill
  \begin{minipage}{0.49\textwidth}
    \centering
    \includegraphics[width=\linewidth]{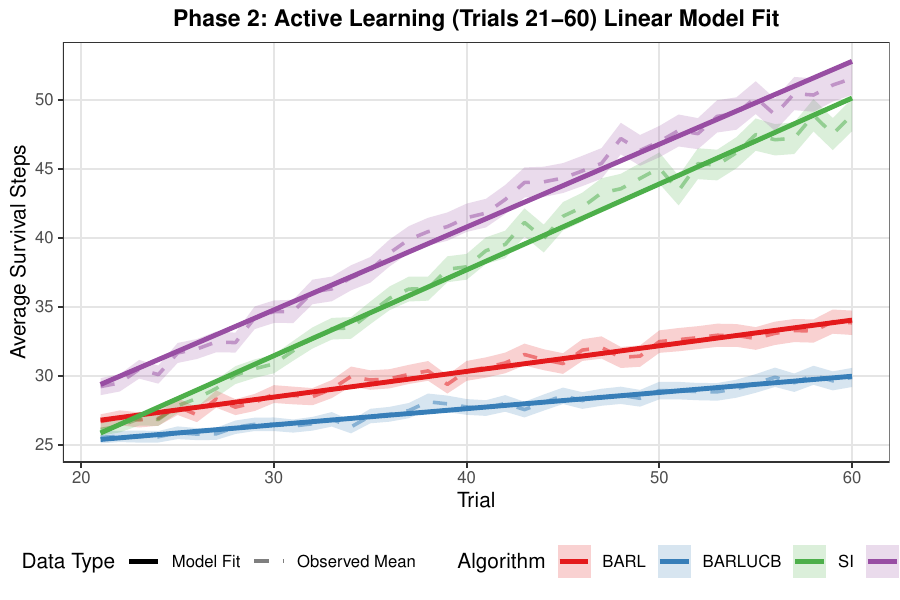}
    \captionof{figure}{Phase 2: Active Learning (Trials 21–60). Observed means with LME model fits and 95\% CIs.}
    \label{fig:phase2}
  \end{minipage}
\end{figure}

Figure~\ref{fig:explorative} shows average survival curves across 120 trials in our main simulations where the likelihood model needed to be learned. These results highlighted clear differences in learning trajectories between algorithms. Most notably, SL improved in performance more quickly than each of the other algorithms and maintained a slight but consistent advantage in later trials.

\subsubsection*{Phase 1: Ramp-up (Trials 1–20)}

During the initial 20 trial phase (Figure~\ref{fig:phase1}), the LME model revealed significant main effects of Trial ($F(1, 157993) = 280.15$, $p < .0001$) and Algorithm ($F(3, 157993) = 122.01$, $p < .0001$), as well as a robust Trial × Algorithm interaction ($F(3, 157993) = 186.59$, $p < .0001$), indicating differential early learning rates.

Post-hoc contrasts (Table~\ref{tab:phase1_slopes_20}) confirmed these results were accounted for by faster learning in SL than in all other algorithms. Namely, the slope for SL ($0.266 \pm 0.007$ SE) exceeded that of SI ($0.107 \pm 0.007$), BARL ($0.113 \pm 0.007$), and BARL-UCB ($0.050 \pm 0.007$), with all pairwise comparisons significant at $p < .0001$.

By the end of this phase (Table~\ref{tab:phase1_emm20}), SL had also already established a lead in average survival (EMM = $27.42 \pm 0.08$), outperforming SI ($24.80 \pm 0.08$), BARL ($26.18 \pm 0.08$), and BARL-UCB ($24.85 \pm 0.08$), with all differences significant ($p < .0001$). 

One other notable observation was that BARL algorithms performed slightly better in initial trials, particularly in trials 1-5. This was most likely due to the fact that SI/SL algorithms prioritized the hill state to a greater degree. In earlier trials, while this benefited learning (and thus led to better performance in later trials) it was not necessarily better for survival when starting with no prior knowledge about the environment. 

\subsubsection*{Phase 2: Active Learning (Trials 21–60)}

In the subsequent phase (Figure~\ref{fig:phase2}), the Trial × Algorithm interaction remained highly significant ($F(3, 317993) = 2183.01$, $p < .0001$), reflecting continued divergence in learning trajectories.

Both SL and SI displayed steep positive slopes, with SI marginally outpacing SL ($0.623 \pm 0.006$ vs. $0.601 \pm 0.006$; difference $= 0.022 \pm 0.008$, $z = 2.72$, $p = 0.0334$). This was likely due to the early success of SL, causing it to have a higher base performance from the start of phase 2 (i.e., the slope for SI reflected the fact that it was catching up to performance levels of SL). However, the difference between these algorithms was small in magnitude, and both far outperformed BARL and BARL-UCB (Table~\ref{tab:phase2_slopes_60}).

Despite the slightly higher slope shown by SI in this phase, SL nonetheless maintained and expanded its performance lead. At Trial 40 (midpoint of this phase), the EMM for SL ($40.78 \pm 0.08$) remained significantly above that of SI ($37.68 \pm 0.08$), with a difference of $3.10 \pm 0.09$ ($z = 33.18$, $p < .0001$), confirming the enduring benefit of faster initial learning in SL.

\begin{table}[htbp]
\centering
\caption{Estimated Mean Survival (EMM $\pm SE$) at Trial 40 (Midpoint of Active Learning Phase, Trials 21--60). Contrasts are SL vs. Other Algorithms.}
\label{tab:phase2_emm40}
\begin{tabular}{lcccccc}
\toprule
Algorithm & EMM Survival & $SE_{\text{EMM}}$ & $\Delta_{\text{EMM}}$ (SL - Comparison)& $SE_{\text{contrast}}$ & $z$ & $p$-value \\
\midrule
SL    & 40.78 & 0.08 & --    & --    & --     & -- \\
SI    & 37.68 & 0.08 & 3.10  & 0.09  & 33.18  & $< .0001$ \\
BARL& 30.31 & 0.08 & 10.47 & 0.09  & 112.03 & $< .0001$ \\
BARL-UCB& 27.61 & 0.08 & 13.17 & 0.09  & 140.93 & $< .0001$ \\
\bottomrule
\end{tabular}
\caption*{\footnotesize Note: EMMs derived from LME model for trials 21--60. $SE_{\text{EMM}}$ is the associated standard error. $\Delta_{\text{EMM}}$ is the estimated difference between SL and the comparison algorithm in question. $SE_{\text{contrast}}$ is the standard error of this difference.}
\end{table}

\subsection{Additional Behavioral Patterns}
Single-trial simulations of the two ActInf algorithms (SI and SL) also revealed interesting patterns of behavior and dependence on choice of preference precision. As this precision effectively down-weights exploratory drives in the EFE, we found that it controlled the total number of time steps the agent spent at the hill (i.e., resolving uncertainty). For these single-trial simulations, we also examined cases in which resource locations were known but season transitions were not, as we found they provided additional insights regarding parameter dependence. For example, Figure \ref{fig:example1} (\textbf{C}) shows a case in which the preference precision was high ($c = 1$), the likelihood function was known, but the transition function was unknown. In this case, both SI and SL agents, despite lacking information about the current context, initially ignored the hill and attempted to infer the current season by visiting the resource positions they knew were associated with a particular context (due to having a precise likelihood model). This was because the epistemic term had a proportionally lower impact when compared to the agent's preferences. Thus, these agents' behavior was driven by the drive to meet their multi-objective preferences, rather than to seek information in the form of large posterior updates to their beliefs about hidden states. This is in line with the classic \textit{risk-seeking} behavior previously described in the ActInf literature \citep{smith_tutorial}. For the SI algorithm, similar behavior was seen when the epistemic term was omitted, regardless of the preference precision.

Recall that, when full knowledge of the environment (transition and likelihood functions) was available, all algorithms, showed greater similarity in performance. When analyzed at the single-trial level, each would often initially move to the hill before proceeding to a resource location (with SL doing so more frequently due to its additional epistemic drive, as mentioned above). This highlights a core similarity between ActInf and BARL. That is, both approaches are Bayes-optimal with respect to their prior beliefs, meaning that, given an initial belief state and a mechanism to calculate the value of some subset of additional belief states (for example, all belief states reachable from the initial belief state up to some horizon, as is the case in these implementations), each agent will optimally calculate the value of each of these belief states. Given a deterministic and greedy policy construction procedure, an optimal policy will then be chosen that maximizes expected value. The major difference then arises when active learning is required to resolve uncertainty about contingencies within the environment.

Another important consideration is that the accuracy with which each algorithm calculates the value of belief states is entirely predicated on the initial belief state. Thus, if the initial belief state is inaccurate, the calculation and evaluation of subsequent belief states will also be inaccurate. Therefore, in simulations where the transition model was known, but the initial context was unknown, the agent knew that transitions were relatively static (95\% chance to remain in the same context and 5\% chance to transition to the next context) and so often viewed visiting the hill as optimal - as it was the state that would most precisely update its belief about the current season. Due to the nature of the counterfactual trajectory planning these agents implement, they search through all possible belief trajectories up to the set planning horizon, and thus calculate, ahead of time, the optimal set of subsequent actions for whatever observation the hill state provides. Planning trajectories then calculate that the hill will provide precise context information, and for each observation the hill could provide, the optimal trajectory following on from that time point is calculated. These belief trajectories thus have high precision compared to those that do not include the hill.

As briefly mentioned above, some initial exploratory analyses also showed interesting ways in which behavioral patterns were influenced by time limits to reach each resource. For example, if one increased these time limits compared to the main simulations shown above (i.e., to 30 time steps without reaching a resource), all agents would initially ignore the hill and simply guess at the context. This was because the agent did not believe it would incur the penalty of reaching the time limit. Thus, it would lose little by guessing at contexts, even if its guesses were wrong. In these scenarios, the agent would often initially move toward resources based on a guess about the context, and only move to the hill if it believed that subsequent guessing would have a higher chance of incurring death. Mathematically, this is due to the agent precisely following the actions it believed would yield the largest return in the expectation, as is the case with all Bayes-optimal algorithms.

In combination, the analyses above highlight ways in which fixed parameters (i.e., preference precision, initial belief states, expected resource time limits, planning horizon) influence decision-making in specific ways. This opens up the possibility of using such models in future studies to capture (and mechanistically explain) individual differences in human cognition and behavior, and potentially their biological basis. This therefore represents one important research direction going forward.

\subsection{Summary}
\label{subsec:conclusion-results}

The experimental results presented in this section provide clear insights into the comparative performance and adaptability of SL, SI, and BARL under different levels of environmental uncertainty. SL consistently outperformed SI and BARL across varying conditions in our novel testing environment, particularly in versions requiring long-term planning that took information value into consideration.

The inclusion of a UCB-style exploration bonus in BARL improved its adaptability, but remained less effective than the intrinsic epistemic and novelty-driven exploration shown by SL. Namely, while the UCB term enabled more directed likelihood learning, it did not fully replicate the structured, hierarchical search mechanisms and state exploration drive inherent in ActInf-based methods.

Tree-search depth and memoization significantly impacted performance trade-offs, particularly for SI. While deeper tree search improved long-term planning, computational costs increase nonlinearly. Memoization appeared to offer a practical solution by caching intermediate search results, but its high memory requirements would need to be carefully managed in large-scale applications.

\section{Discussion}
\label{sec:discussion}

This study aimed to (1) compare the performance of Sophisticated Inference (SI) with Bayes-adaptive reinforcement learning methods (BARL), and (2) introduce and evaluate Sophisticated Learning (SL), an extension of SI that integrates active learning into recursive planning. Our simulations, conducted in a novel, biologically inspired grid-world task, provide key insights into the behavior and comparative strengths and weaknesses of these algorithms.

\subsection{Key Findings and General Contribution}

SL outperformed SI  and BARL (with and without an upper confidence bound [UCB] heuristic promoting directed exploration) across all simulations where model-learning was required. Here, performance was measured by the number of time steps survived per trial, which inherently relies on an agent's ability to learn an accurate model. Trial-by-trial variance was observed due to the inherent difficulty of the task, but on average, SL demonstrated superior performance. This reflected its novel ability to strategically revisit states based on anticipated future observations, balancing exploration and exploitation over multiple future time steps.

Unlike BARL algorithms, which focus on maximizing expected cumulative rewards, SL leverages expected information gain to guide behavior. In particular, the SL agent used forward-looking strategies, simulating how future observations would update its beliefs about earlier states and state-outcome mappings. A notable pattern emerged: upon discovering a resource, the SL agent often revisited a state that disambiguated the current context (the hill). This behavior exemplifies the ability of SL to link observations across time to improve its contextual understanding, a feature absent from the other algorithms.

Putting a psychological gloss on this mechanism, an agent employing this algorithm might engage in the following thought process:

\begin{quote}
\textit{I have now discovered a state where a food resource is located. I am unsure of what season I am in at this point, but if I were to move from here and visit the hill state, it would tell me what season I am in. Then, given my transition model, I would be able to work backwards and retrospectively figure out what season I was most likely in when I was at the position of the food. Although not maximally precise, visiting the hill would allow me to do this with more precision than moving to some other state that would not improve my knowledge of what context I am in. This, in turn, would allow me to assign a context to that specific position of the food, which I can exploit in the future.}
\end{quote}

This highlights the ability of SL to anticipate how future observations will update posterior beliefs about the past, thereby optimizing exploration toward states that improve contextual understanding. In this way, SL offers a more strategic, nuanced form of directed exploration, focusing not just on visiting new states, but on those expected to improve current beliefs about past rewards.

\subsection{Exploration Strategies: SI vs. SL vs. RL}

In line with the description above, the advantages of SL over SI therefore appear to be due to its capacity for backward counterfactual reasoning — anticipating the benefit of future observations in refining past beliefs. While SI displayed strong performance through a more classic form of directed exploration (e.g., seeking unvisited states), it lacked the ability to leverage beliefs about how future observations could be strategically used to update contextual understanding of previous observations.

While the performance differences between SL and SI were relatively small, the common elements of their exploratory strategies led to much larger performance advantages over BARL. Adding UCB-based directed exploration to  BARL also did not improve its relative performance. Instead, it led to over-exploration of states with low epistemic affordances. This ultimately reduced efficiency, as the hill state was not weighted any differently in its epistemic evaluation than any other unvisited state. These findings highlight the difference between intrinsic curiosity about state-outcome mappings in UCB, the further curiosity about current contextual states in SI, and the strategic, goal-directed exploration shown by SL.

\subsection{Mechanisms of suboptimal performance}

Despite the high performance of SL relative to comparison agents, it nonetheless frequently failed to converge onto optimal policies, leading to high variance in performance across trial sequences. Understanding these failures highlighted core challenges in learning under uncertainty, but also illuminated issues that may be less general and contingent on the specific environment and implementation being considered.

In some of our supplementary analyses (\ref{fig:divergence_example}), a common failure mode we observed arose from early epistemic commitments in SL. In particular,  while SL agents use recursive planning to project belief updates forward, this mechanism is only as reliable as the evidence received. During early learning, incorrect associations between contexts and resource locations (e.g., from low-probability observations) can become entrenched, as Dirichlet counts accumulate in favor of incorrect likelihoods. Once an incorrect model is strongly reinforced, the agent tends to commit to poor policies - such as moving toward an expected resource that is, in fact, absent. Because the agent’s own model can be misleading in such cases, and survival windows are limited, these trajectories often preclude the opportunity to learn a more accurate model. This form of \textit{self-confirming bias} is particularly problematic in sparse-reward or high-penalty environments, such as the one tested here.

This effect is visible in \textit{appendix }Figure~\ref{fig:divergence_example}, where belief distributions in Season 3 diverge over time from the true Sleep resource locations. Rather than improving, the model degrades due to a reinforcing loop between inaccurate inference and maladaptive behavior. Notably, this issue is not due to insufficient planning depth, but rather to incorrect parameter learning that remains uncorrected.

Another interesting aspect of this stems from the \textit{backward smoothing} mechanism used within SL (see Section~\ref{sec:SL} and Appendix \ref{app:sl_tree_search}). This mechanism is intended to revise posterior beliefs over states from earlier time steps in light of new observations. In principle, this should allow agents to forget and/or correct past inferences and improve parameter learning, even after delayed evidence is received. However, the divergence patterns shown in Figure~\ref{fig:divergence_example} suggest that these mechanisms are not always sufficient. Once strong - but incorrect - beliefs are established, even recursive smoothing may be unable to dislodge them, particularly in regimes with ambiguous feedback.

However, it should be noted that this failure mode is not intrinsic to SL in particular, but rather reflects the interaction between priors, environmental structure, and chosen hyperparameters (e.g., learning rate, planning depth, initial Dirichlet counts) that can be present in any Bayesian agent. The agent’s initial uncertainty, the rate at which beliefs are updated, and the asymmetry of risks across contexts all shape learning trajectories. SL, like any Bayesian learner, is sensitive to its initial conditions. Thus, these observed failures - while instructive - should not be over-interpreted as a major limitations of the SL algorithm \textit{per se}.

Indeed, when the agent is provided with a correct generative model (Section~\ref{horiz_analysis_res}), performance improves markedly, confirming that accurate beliefs are the primary bottleneck to adaptive behavior. Additionally, the KL-divergence measures of learning in each context shown in \ref{fig:divergence_example} revealed that certain resources or seasons are harder to learn, most likely due to their statistical properties or positional inaccessibility.

Future work could explore mechanisms for \textit{meta-inference} - enabling the agent to represent and revise its confidence in its own beliefs - or explore the use of other forgetting strategies, such as `belief resets' upon persistent mismatch between predicted and real observations. 

\subsection{Limitations and Future Directions}

While SL demonstrates clear advantages in this specific environment, several caveats should be considered. First, the grid-world environment was chosen to test the expected advantages of SL; thus, future research will be needed to determine the extent to which the benefits of SL generalize to other environments. Here, we would expect SL to excel in tasks requiring deep planning and strategic exploration, but its relative performance across a wide variety of more traditional RL-style benchmarks remains uncertain.

Another consideration involves the optimization of parameter values. For example, the optimal value for preference precision in SL may vary across tasks. Some tuning is likely to be required to balance the epistemic and novelty terms in the EFE for different problems. This sensitivity to parameterization will be an important practical factor when applying SL in diverse environments.

A further limitation concerns computational efficiency. Like other ActInf algorithms, SL relies on recursive tree search, which can become computationally expensive in real-world environments. Scaling SL to such domains will likely require the integration of other heuristic approaches, more efficient pruning techniques, or other machine learning approximations. Future research should therefore focus on developing methods to enhance the scalability of SL while maintaining its strategic advantages.

\section{Conclusion}
\label{sec:conclusion}

In this study, we used a challenging, dynamic environment requiring complex planning and strategic information-seeking to compare Active Inference and Bayesian reinforcement learning algorithms. We first showed that a recent `Sophisticated Inference' algorithm within the Active Inference framework outperformed Bayesian reinforcement learning in this environment (both with and without the addition of a common directed exploration term). Second, we proposed and tested a novel `Sophisticated Learning' algorithm—combining insights from Sophisticated Inference and Bayesian Reinforcement Learning—and demonstrated further advantages it may offer. This algorithm demonstrated greater performance than either of the other algorithms tested. It also exhibited qualitatively distinct, strategic patterns of behavior in which it gathered information to improve its understanding of past observations. The associated backward reasoning strategies employed by Sophisticated Learning represent a novel advance in simulating intelligent agent behavior. 

These promising results suggest that Sophisticated Learning may offer new insights into both machine learning and cognitive science. Future work should assess the generalizability of the strategies that emerge from this algorithm in other machine learning contexts and investigate whether it might capture unique patterns observed in animal and human behavior, contributing to ongoing research in cognitive and computational neuroscience.

\section*{Acknowledgment}
SJG is supported by funding from the Oppenheimer Memorial Trust. EAB is supported by the Dutch Research Council (NWO), grant number 019.223SG.002. R.S. is supported by the Laureate Institute for Brain Research. Computations were performed, in part, using facilities provided by the University of Cape Town’s ICTS High Performance Computing team: hex.uct.ac.za. Travel and additional compute for this work has been funded by Conscium Ltd.

\section*{Code and Data Availability}
All code for reproducing and customizing the simulations reported in this manuscript can be found at: \url{https://github.com/sgrimbly/Sophisticated-Learning}.

% \bibliographystyle{apalike}
% \bibliography{references.bib}

% Bibliography
% \bibliographystyle{plainnat}
\bibliographystyle{elsarticle-harv}
\bibliography{references}

\section*{Appendix}
This appendix offers more detailed explanations of the algorithms and experimental environment used in this study, as well as additional analyses supporting findings reported in the main text.

% \subsection{Variational Inference}
% \colorbox{yellow}{Necessary?}

\subsection{Detailed Algorithmic Specifications}
\label{app:algorithms}

In this section, we elaborate on the core algorithms discussed in the main paper. While the main text provides an overview of Active Inference (ActInf), Sophisticated Inference (SI), Bayes-Adaptive Reinforcement Learning (BARL), and our proposed Sophisticated Learning (SL) algorithm (as discussed in Sections~\ref{subsec:SI} and \ref{sec:SL}), the following subsections offer detailed pseudocode. This level of detail is provided for reproducibility and for readers interested in the specific computational steps involved in the recursive tree search and learning mechanisms that were utilized.

% \colorbox{yellow}{To include:}
% \begin{enumerate}
%     \item Discussion about robustness tests/grid experiments   
%     \item Chosen baselines - BA, BAUCB details
%     \item Specific algorithm details and quirks
% \end{enumerate}

\subsubsection{Sophisticated Inference (SI) Tree Search}
\label{app:si_tree_search}

The SI algorithm, introduced in Section~\ref{subsec:SI} of the main text, extends ActInf by employing a recursive tree search to evaluate policies based on their Expected Free Energy (EFE). This approach dynamically constructs policies rather than relying on a pre-specified policy space. The EFE calculation, as formulated, balances exploration (information gain about states and parameters) and exploitation (receipt of preferred observations). Algorithm~\ref{alg:si_tree_search_appendix} details the associated forward tree search mechanism. 

The SI search algorithm forms the basis from which SL was developed. Understanding this search process is therefore crucial for appreciating the extensions introduced in SL. In particular, it underlies the way SL incorporates active learning by simulating model parameter updates within the associated recursive structure (see Section~\ref{sec:SL} in the main text).  Both algorithms also employ the same pruning mechanisms for tractability. These are implicitly handled within the `ViableActions` and `LikelyStates` components of the following pseudocode, which present the SI tree search process in detail.

\begin{algorithm}[h]
\caption{Sophisticated Inference Tree Search}\label{alg:si_tree_search_appendix}
\begin{algorithmic}
\Function{ForwardTreeSearch}{$posterior$, $A$, $a$, $b$, $observation$, $t_{resources}$, $t$, true\_t, $T$, $W$, $h$}

\If{$t$ $>$ true\_t}
    \State $posterior$ = CalculatePosterior($posterior$, $b$, $a$, $observations$)
    \State $t_{resources}$ $\gets$ NewTimeSinceResources($posterior(t)$, resource\_locations)
\EndIf
\For{each action}
    \State Q($action$) = b($posterior(t)$, $action$) 
    \State $predicted\_observations$ $\gets$ a(Q($action$)
    \State $G(action)$ $\gets$ ExpectedFreeEnergy($predicted\_observations$, $t_{resources}$, $W$)
\EndFor
\If{t $<$ $h$}
    \State $actions$ $\gets$ ViableActions($G$)
    \For{$action$ in $actions$}
        \State $posterior(t+1)$ $\gets$ Q$(action$)
        \State $states$ $\gets$ Q($action$)
        \State $states$ $\gets$ LikelyStates($states$)
        \For{$state$ in $states$}
            \State $observation$ $\gets$ SampleObservation(state, $a$)
            \State [$G$, $posterior$] $\gets$ ForwardTreeSearch($posterior$, $a$, $b$, \State $observation$, $t_{resources}$, $t+1$, true\_t, $T$, $W$, $h$)
            \State $S$ $\gets$ softmax($G$) * $G$
            \State K($state$) $\gets$ $S$
        \EndFor
        \State $G$(action) $\gets$ K($states$)*$states$
    \EndFor
\EndIf \\
    \Return [$G$, $posterior$]
\EndFunction
\end{algorithmic}
\end{algorithm}

\begin{figure}[h]
\centering
\begin{tikzpicture}[
    level distance=2.8cm,
    sibling distance=10cm,
    edge from parent path={(\tikzparentnode) -- (\tikzchildnode)},
    every node/.style={align=center},
    planning/.style={rectangle, rounded corners, draw=black, fill=blue!10, 
        text width=2.8cm, minimum height=0.8cm, inner sep=4pt},
    decision/.style={diamond, draw=black, fill=green!10, 
        text width=2.5cm, inner sep=2pt, aspect=2.5},
    action/.style={rectangle, draw=black, fill=orange!10, 
        text width=2.8cm, minimum height=0.8cm},
    outcome/.style={rectangle, rounded corners, draw=black, fill=yellow!10, 
        text width=2.8cm},
    arrow/.style={->, >=stealth, thick},
    note/.style={text width=2.5cm, font=\small\itshape, fill=none},
    energy/.style={rectangle, draw=red!60, dashed, fill=red!5, 
        text width=2.5cm, font=\small, inner sep=2pt}
]

% Root level
\node[decision] (time) at (0,0) {Has agent moved\\in real time?};

% Left branch with adjusted spacing
\node[planning] (update) at (-5,-2) {Update beliefs};
\node[action] (actions) at (-5,-4) {Evaluate Future Actions Using EFE};
\node[decision] (horizon) at (-5,-6.5) {Within \\horizon?};
\node[planning] (simulate) at (-7,-9) {Simulate future\\outcomes};
\node[action] (recursive) at (-7,-11) {Continue search\\with updates};
\node[outcome] (return) at (-7,-13) {Return optimal\\action};
\node[outcome] (estimate) at (-3,-9) {Return current\\estimate};

% Right branch
\node[action] (evaluate) at (5,-2) {Evaluate Future Actions\\ Using EFE};
\node[decision] (horizon2) at (5,-4.5) {Within horizon?};
\node[planning] (continue) at (3,-6.5) {Continue search};
\node[outcome] (return2) at (6.5,-6.5) {Return\\evaluation};

% Connections
\draw[arrow] (time) -- node[left] {Yes} (update);
\draw[arrow] (time) -- node[right] {No} (evaluate);
\draw[arrow] (update) -- (actions);
\draw[arrow] (actions) -- (horizon);
\draw[arrow] (horizon) -- node[left] {Yes} (simulate);
\draw[arrow] (horizon) -- node[right] {No} (estimate);
\draw[arrow] (simulate) -- (recursive);
\draw[arrow] (recursive) -- (return);
\draw[arrow] (evaluate) -- (horizon2);
\draw[arrow] (horizon2) -- node[left] {Yes} (continue);
\draw[arrow] (horizon2) -- node[right] {No} (return2);

\end{tikzpicture}
\caption{Flowchart representation of the Sophisticated Inference Tree Search algorithm in ActInf. This tree search process balances exploration (information gain about states and parameters) with goal-seeking behavior (reward value) by recursively evaluating actions using EFE. The algorithm begins by checking if the agent has moved in real time. If it has, it updates beliefs via Variational Free Energy (VFE) minimization and then evaluates possible actions based on EFE. If the current time step is within the planning horizon, the system simulates likely future states and checks whether past evaluations exist in short-term memory. If the state has been previously explored, stored evaluations are retrieved to optimize search efficiency. If no prior evaluation exists, the tree search recursively expands, exploring deeper future states until the optimal action is determined. When recursion completes, the best action sequence is returned. If the agent has not moved in real time, actions are evaluated externally, and search continues within the planning horizon. This process ensures the model dynamically balances exploration and exploitation in complex decision-making scenarios.}
\label{fig:si-tree-search-flowchart}
\end{figure}

\begin{figure}[h]
\centering
\begin{tikzpicture}[
    level distance=2.8cm,
    sibling distance=8cm,
    edge from parent path={(\tikzparentnode) -- (\tikzchildnode)},
    every node/.style={align=center},
    decision/.style={diamond, draw=black, fill=green!10, 
        text width=3cm, inner sep=2pt, aspect=2.5},
    action/.style={rectangle, draw=black, fill=orange!10, 
        text width=3cm, minimum height=0.8cm},
    outcome/.style={rectangle, rounded corners, draw=black, fill=yellow!10, 
        text width=3cm},
    arrow/.style={->, >=stealth, thick}
]

% Root Decision: Is within planning horizon?
\node[decision] (horizon) at (0,0) {Within planning horizon?};

% Left branch (Tree Search Expands)
\node[action] (actions) at (-5,-3) {Evaluate actions using EFE};
\node[decision] (state) at (-5,-6) {Select likely states};
\node[decision] (memory) at (-5,-9) {State previously evaluated?};
\node[action] (recursive) at (-5,-12) {Expand tree recursively};
\node[outcome] (return) at (-5,-15) {Return best action};

% Memory retrieval path
\node[outcome] (memory_retrieve) at (-1,-12) {Retrieve stored value};

% Right branch (Terminate Search)
\node[outcome] (estimate) at (3,-3) {Return current estimate};

% Connections
\draw[arrow] (horizon) -- node[left] {Yes} (actions);
\draw[arrow] (horizon) -- node[right] {No} (estimate);
\draw[arrow] (actions) -- (state);
\draw[arrow] (state) -- (memory);
\draw[arrow] (memory) -- node[left] {No} (recursive);
\draw[arrow] (memory) -- node[right] {Yes} (memory_retrieve);
\draw[arrow] (recursive) -- (return);
\draw[arrow] (memory_retrieve) -- (return);

\end{tikzpicture}
\caption{Tree search logic for Sophisticated Inference (SI). This tree search evaluates actions using EFE, selects likely states, and checks short-term memory for past evaluations. If a state has been visited previously, its stored evaluation is retrieved to optimize search efficiency. Otherwise, the search expands recursively until the optimal action is found. The recursion terminates when the planning horizon is reached.}
\label{fig:si-tree-search}
\end{figure}

\subsubsection{Bayes-Adaptive Reinforcement Learning (BARL) Tree Search}
\label{app:barl_tree_search}

For comparative analyses, we implemented the BARL method \citep{Ross2007}. The planning structure for this agent also utilized a recursive tree search, analogous to SI. However, BARL aims to maximize an explicit reward function integrated over the belief space of model parameters, rather than minimizing EFE. Algorithm~\ref{alg:barl_tree_search_appendix} provides the pseudocode.

The key distinction here is use of an explicit reward function (based on the time since each resource) with no intrinsic exploration component and direct updating of model parameters (concentration parameters $a, b$) within each recursive step of the planning phase. In some simulations, an Upper Confidence Bound (UCB) term was also incorporated into the `Reward` calculation step for BARL to encourage directed exploration by adding a bonus for less visited state-action pairs, using counts $N_t$. The addition of UCB provided a closer comparison to the ActInf algorithms, as performance within the testing environment benefited from their directed exploratory behavior.

The following pseudo-code depiction of the BARL algorithm shows UCB included, with $Nt$ representing the number of visits to each state up until the current time point.

\begin{algorithm}[h]
\caption{Bayes-Adaptive Tree Search}\label{alg:barl_tree_search_appendix}
\begin{algorithmic}
\Function{ForwardTreeSearch}{$posterior$, $A$, $a$, $b$, $observation$, $t_{resources}$, $t$, true\_t, $T$, $h$, $N_t$}

\If{$t$ $>$ true\_t}
    \State $posterior$ = CalculatePosterior($posterior$, $b$, $a$, $observations$)
    \State $t_{resources}$ $\gets$ NewTimeSinceResources($posterior(t)$, resource\_locations)
\EndIf
\State $a, b$ $\gets$ UpdateConcentrationParameters($posterior$, $a$, $b$ o$bservation$)
\For{each $action$}
    \State Q($action$) = b($posterior(t)$, $action$) 
    \State predicted\_observations $\gets$ a(Q($action$)
    \State R($action$) $\gets$ Reward(predicted\_observations, $t_{resources}$, $N_t$, Q($action$))
\EndFor
\If{t $<$ $h$}
    \State actions $\gets$ ViableActions(R)
    \For{action in actions}
        \State $posterior(t+1)$ $\gets$ Q($action$)
        \State $states$ $\gets$ Q($action$)
        \State $states$ $\gets$ LikelyStates(states)
        \For{$state$ in $states$}
            \State $N_t$ $\gets$ UpdateStateVisits($N_t$, $state$)
            \State $observation$ $\gets$ SampleObservation($state$, $a$)
            \State [R, $posterior$] $\gets$ ForwardTreeSearch($posterior$, $a$, $b$, \State $observation$, $t_{resources}$, $t+1$, true\_t, $T$, $h$, $N_t$)
            \State $S$ $\gets$ softmax($R$) * $R$
            \State K($state$) $\gets$ $S$
        \EndFor
        \State R($action$) $\gets$ K($states$)*$states$
    \EndFor
\EndIf \\
    \Return [R, $posterior$]
\EndFunction
\end{algorithmic}
\end{algorithm}

\subsubsection{Sophisticated Learning (SL) Tree Search and Backwards Smoothing}
\label{app:sl_tree_search}

The SL algorithm detailed in Section~\ref{sec:SL} of the main text extends SI by integrating active learning into the planning process using insights from BARL. This is achieved by allowing the agent to counterfactually reason about how its model parameters would evolve based on future hypothetical observations and then refining its understanding of past states through backward smoothing.

\paragraph{Backward Smoothing Mechanism}
Backward smoothing was conceptually introduced in Section~\ref{subsec:SI} and depicted in the flowchart within Figure~\ref{fig:backwards_smoothing}. This mechanism allows the agent to retrospectively adjust its posterior beliefs over states at previous time steps within a simulated trajectory, given a sequence of actions and subsequent (hypothetical) observations. Algorithm~\ref{alg:backwards_smoothing_appendix} details this process.

\begin{algorithm}[H]
\caption{Backwards Smoothing}\label{alg:backwards_smoothing_appendix}
\begin{algorithmic}[1] % [1] enables line numbering
\Function{BackwardSmoothing}{$\textit{posterior}$, $a$, $b$, $\textit{observation}$, \textit{action\_history}, $t$, $\tau$}
    \State $L \gets \textit{posterior}$
    \State $p \gets 1$
    \For{$\textit{timestep} = t + 1$ to $\tau$}
        \State $p \gets b(\textit{action\_history}(timestep - 1)) \times p$
        \For{$\textit{state}$ in $L$}
            \State $L(\textit{state}) \gets L(\textit{state}) \times \textit{observation(timestep)} \times a \times p(\textit{state})$
        \EndFor
    \EndFor
    \Statex
    \Return $\textproc{normalise}(L)$
\EndFunction
\end{algorithmic}
\end{algorithm}

This smoothing procedure is particularly beneficial for learning the observation model ($A$), and potentially for learning a transition model ($B$) as well (although we do not explicitly test transition learning here). By refining the posterior probability of having been in a particular state $s_{t'}$ given observations up to $s_t$ (where $t > t'$), the agent can make more accurate updates to its Dirichlet concentration parameters. This allows observed outcomes to be better assigned to past grid positions within their respective latent contexts, especially in partially observable environments.

\paragraph{Full Sophisticated Learning (SL) Tree Search}
The full SL tree search algorithm synergistically combines recursive EFE-based planning in SI with further elements that facilitate active learning. As mentioned above, these elements include simulated updates to Dirichlet concentration parameters ($a, b$), which represent beliefs about the observation and transition models, respectively, and backward smoothing of posterior state beliefs. Algorithm~\ref{alg:sl_tree_search_appendix} outlines this process in full.

\begin{algorithm}[h]
\caption{Sophisticated Learning Tree Search}\label{alg:sl_tree_search_appendix}
\begin{algorithmic}
\Function{ForwardTreeSearch}{$posterior$, $A$, $a$, $b$, $observation$, $tResources$, $t$, $trueT$, $T$, $h$, $backwardsHorizon$, $actionHistory$, $resourceLocations$}

    \Comment{Update posterior and resource timing if past the ground truth time}
    \If{$t > trueT$}
        \State $posterior \gets$ CalculatePosterior($posterior$, $b$, $a$, $observations$)
        \State $tResources \gets$ NewTimeSinceResources($posterior(t)$, $resourceLocations$)
    \EndIf

    \Comment{Set the starting time for backward smoothing}
    \State $startTime \gets \max(1, t - backwardsHorizon)$

    \Comment{Initialize novelty tracking}
    \State $novelty \gets 0$
    \State $aPrior, bPrior \gets a, b$

    \Comment{Perform backward smoothing to update concentration parameters}
    \For{$\tau = startTime$ to $t$}
        \State $L \gets$ BackwardSmoothing($posterior$, $a$, $observations$, $actionHistory$, $t$, $\tau$)
        \State $a, b \gets$ UpdateConcentrationParameters($L$, $b$, $observations$, $a$)
    \EndFor

    \Comment{Compute novelty weight based on parameter changes}
    \State $W \gets$ CalculateNovelty($a, b, aPrior, bPrior$)

    \Comment{Evaluate expected free energy (EFE) for each action}
    \For{each $action$ in $A$}
        \State $Q(action) \gets b(posterior(t), action)$
        \State $predictedObservations \gets a(Q(action))$
        \State $G(action) \gets$ ExpectedFreeEnergy($predictedObservations$, $tResources$, $N_t$, $Q(action)$)
    \EndFor

    \Comment{Perform recursive tree search if time step is within horizon}
    \If{$t < h$}
        \State $viableActions \gets$ ViableActions($G$)
        \For{$action$ in $viableActions$}
            \State $posterior(t+1) \gets Q(action)$
            \State $likelyStates \gets$ LikelyStates($Q(action)$)

            \For{$state$ in $likelyStates$}
                \State $observation$ $\gets$ SampleObservation($state$, $a$)
                \State [$G$, $posterior$] $\gets$ ForwardTreeSearch($posterior$, $a$, $b$, 
                \Statex \hspace{40mm} $observation$, $tResources$, $t+1$, $trueT$, $T$, 
                \Statex \hspace{40mm} $h$, $backwardsHorizon$, $actionHistory$, $resourceLocations$)
                \State $S$ $\gets$ softmax($G$) * $G$
                \State $K(state)$ $\gets$ S
            \EndFor

            \State $G(action) \gets$ Sum($K(likelyStates) \cdot likelyStates$)
        \EndFor
    \EndIf

    \Return [$G$, $posterior$]
\EndFunction
\end{algorithmic}
\end{algorithm}

The integration of `BackwardSmoothing' and `UpdateConcentrationParameters' within each recursive step (and the subsequent calculation of a '$W_{novelty}$' term derived from the expected change in model parameters) is what fundamentally distinguishes SL from SI. This allows SL to explicitly plan actions that are expected to yield high `novelty' – i.e., actions that will maximally reduce uncertainty about the model parameters themselves. This forward simulation of possible learning trajectories enables the agent to make proactively self-improving decisions, going beyond simple state exploration to engage in true parameter exploration. The psychological interpretation of this process is discussed in the main text.

\subsection{Further Results}
Table \ref{tab:phase1_slopes_20} below displays the mean estimated survival slope fit for each algorithm for the first 20 trials within LMEs. These results re-iterate the significant survival advantage SL had in earlier trials, leading to convergence 40\% more quickly than SI. 
\begin{table}[htbp]

\centering
\caption{Learning Rates (Mean Slope $\pm SE$) in Ramp-up Phase (Trials 1--20). Contrasts are SL vs. Other Algorithms.}
\label{tab:phase1_slopes_20}
\begin{tabular}{lcccccc}
\toprule
Algorithm & Mean Slope & $SE_{\text{slope}}$ & $\Delta_{\text{slope}}$ (SL - Comparison)& $SE_{\text{contrast}}$ & $z$ & $p$-value \\
\midrule
SL    & 0.266 & 0.007 & --    & --    & --     & -- \\
SI    & 0.107 & 0.007 & 0.159 & 0.010 & 16.61  & $< .0001$ \\
BARL& 0.113 & 0.007 & 0.152 & 0.010 & 15.93  & $< .0001$ \\
BARL-UCB& 0.050 & 0.007 & 0.216 & 0.010 & 22.58  & $< .0001$ \\
\bottomrule
\end{tabular}
\caption*{\footnotesize Note: Estimated marginal trends (slopes) derived from the LME model for trials 1--20. $SE_{\text{slope}}$ is the associated standard error. $\Delta_{\text{slope}}$ is the estimated difference between SL and the algorithm in question. $SE_{\text{contrast}}$ is the standard error of this difference.}
\end{table}

\begin{table}[htbp]
\centering
\caption{Estimated Mean Survival (EMM $\pm SE$) at Trial 20 (End of Ramp-up Phase). Contrasts are SL vs. Other Algorithms.}
\label{tab:phase1_emm20}
\begin{tabular}{lcccccc}
\toprule
Algorithm & EMM Survival & $SE_{\text{EMM}}$ & $\Delta_{\text{EMM}}$ (SL - Comparison)& $SE_{\text{contrast}}$ & $z$ & $p$-value \\
\midrule
SL    & 27.42 & 0.08 & --    & --    & --     & -- \\
SI    & 24.80 & 0.08 & 2.62  & 0.11  & 24.64  & $< .0001$ \\
BARL& 26.18 & 0.08 & 1.24  & 0.11  & 11.68  & $< .0001$ \\
BARL-UCB& 24.85 & 0.08 & 2.57  & 0.11  & 24.18  & $< .0001$ \\
\bottomrule
\end{tabular}
\caption*{\footnotesize Note: EMMs derived from the LME model for trials 1--20, evaluated at Trial 20. $SE_{\text{EMM}}$ is the association standard error. $\Delta_{\text{EMM}}$ is the estimated difference between SL and the comparison algorithm in question. $SE_{\text{contrast}}$ is the standard error of this difference.}
\end{table}

\begin{table}[htbp]
\centering
\caption{Learning Rates (Mean Slope $\pm SE$) in Active Learning Phase (Trials 21--60). Key contrast is SI vs. SL.}
\label{tab:phase2_slopes_60}
\begin{tabular}{lcccccc}
\toprule
Algorithm & Mean Slope & $SE_{\text{slope}}$ &  $\Delta_{\text{slope}}$ (SL - Comparison)& $SE_{\text{contrast}}$ & $z$ & $p$-value \\
\midrule
SL    & 0.601 & 0.006 & --     & --    & --     & -- \\
SI    & 0.623 & 0.006 & 0.022  & 0.008 & 2.72   & $0.0334$ \\
BARL& 0.186 & 0.006 & -0.415 & 0.008 & -51.29 & $< .0001$ \\
BARL-UCB& 0.118 & 0.006 & -0.483 & 0.008 & -59.75 & $< .0001$ \\
\bottomrule
\end{tabular}
\caption*{\footnotesize Note: Displayes slopes (estimated marginal trends) within the LME model for trials 21--60. The contrast for SI is SI-SL. Other contrasts are Algorithm-SL. Thus, a positive value for SI-SL means SI had a steeper slope. $SE_{\text{slope}}$ is the standard error for each slope. $\Delta_{\text{slope}}$ is the estimated difference (Algorithm-SL). $SE_{\text{contrast}}$ is the standard error of this difference.}
\end{table}

\subsection{Robustness to Grid Reconfiguration}  
\label{subsubsec:grid-experiments}
While our main analyses (Section~\ref{sec:results}) compared SL, SI, and BARL on the specific configuration of the environment detailed in Section~\ref{subsec:envir}, an important consideration remains: how well do these algorithms generalize across changes in this configuration? A potential concern would be if performance differences in one setting might reflect overfitting rather than a generalizable difference between algorithms \citep{Zhou_2022}. Since many algorithms are implicitly tuned to a particular structure, evaluating robustness under reconfiguration of resource locations is a simple way to validate our results.

To explore this, we evaluated all four algorithms on a set of randomized grid configurations. These maintained the overall problem structure but varied the location of each resource in each season as well as the agent’s start position. This allowed us to test whether the performance advantages seen for SL were due to general adaptability or specific features of the original environment.

We also systematically varied one key structural parameter: the maximum placement distance of resources from the hill. This distance, which we term the `Horizon`, is defined by the Chebyshev distance (i.e., the maximum number of steps in any one cardinal or diagonal direction).

We generated configurations with both a `Horizon` of 3 and a `Horizon` of 5. The `Horizon=5` setting allowed resources to be placed anywhere on the 10x10 grid, while the `Horizon=3` setting constrained placement to a more local 7x7 area around the hill. Initial exploration revealed that configurations with the larger `Horizon=5` setting often resulted in less informative learning dynamics, as some resource locations were too distant to be reliably discovered and exploited within the trial period. This aligns with the theoretical expectation that agents with finite planning depths perform better when key locations are within a reasonable proximity.

Consequently, to create a more challenging and meaningful test of adaptive learning, we focused our main robustness analysis on a dedicated set of 15 configurations where the `Horizon` was fixed at 3. This approach ensured that performance differences reflect the efficiency of each algorithm in exploring a structurally coherent, non-trivial environment. Our analysis was based on a complete dataset of 200 simulation seeds for both the SL and SI algorithms across all 15 of these `Horizon=3` configurations, with each seed being evaluated for 120 trials. Table~\ref{tab:random_grid_configs_appendix} details these configurations.

% We generated 10 additional grid layouts. In these layouts, the 10$\times$10 grid size and the hill’s central position (cell 55, corresponding to coordinates (5,6) in a 1-indexed grid) were preserved. Resource positions (food, water, sleep) across the four seasonal contexts, as well as the agent's starting location, were randomised. A key constraint in these generated configurations was that these randomised positions (resources and start) had to be located within a specific `Horizon` from the hill. This `Horizon` defines a square region around the hill and is based on the Chebyshev distance (also known as the $L_\infty$ norm, or max($\left|\Delta\text{row}\right|$,$\left|\Delta\text{col}\right|$)). As detailed in Table~\ref{tab:random_grid_configs_appendix}, some configurations utilised a `Horizon` of 3 grid units, while others used a `Horizon` of 5 grid units, relative to the hill's position.

\begin{table}[h]
    \centering
    \caption{Details of Randomized Grid Configurations}
    \label{tab:random_grid_configs_appendix}
    \begin{tabular}{cccccc}
        \toprule
        \textbf{Grid Size} & \textbf{Hill} & \textbf{Start} & \textbf{Food} & \textbf{Water} & \textbf{Sleep} \\
        \midrule
        $10 \times 10$ & 55 & 35 & (84,24,33,75) & (86,82,44,74) & (22,62,64,66) \\
        $10 \times 10$ & 55 & 78 & (38,84,43,34) & (32,85,57,76) & (56,82,58,87) \\
        $10 \times 10$ & 55 & 22 & (62,38,68,34) & (32,62,75,76) & (23,74,66,67) \\
        $10 \times 10$ & 55 & 67 & (47,83,25,83) & (78,57,38,82) & (27,62,83,22) \\
        $10 \times 10$ & 55 & 23 & (53,42,36,74) & (73,85,76,76) & (35,82,62,88) \\
        % \midrule
        $10 \times 10$ & 55 & 44 & (75,68,82,74) & (66,26,88,23) & (45,28,57,38) \\
        $10 \times 10$ & 55 & 26 & (47,46,73,78) & (27,37,38,66) & (87,44,74,36) \\
        $10 \times 10$ & 55 & 46 & (68,74,26,34) & (25,63,22,65) & (85,28,36,87) \\
        $10 \times 10$ & 55 & 74 & (48,58,33,86) & (46,76,82,52) & (22,68,38,27) \\
        $10 \times 10$ & 55 & 37 & (58,22,63,85) & (86,32,87,77) & (72,78,65,72) \\
        $10 \times 10$ & 55 & 25 & (36,22,67,37) & (68,57,64,85) & (75,53,77,82) \\
        $10 \times 10$ & 55 & 74 & (72,53,22,26) & (42,86,52,43) & (62,75,58,63) \\
        $10 \times 10$ & 55 & 66 & (43,87,46,22) & (26,63,67,52) & (58,35,23,35) \\
        $10 \times 10$ & 55 & 34 & (26,26,37,52) & (24,54,66,54) & (68,75,73,36) \\
        $10 \times 10$ & 55 & 85 & (47,67,75,52) & (52,43,48,27) & (63,46,73,37) \\
        \bottomrule
    \end{tabular}
\end{table}

\subsection{Comparison of SI and SL on Selected Configurations}
\label{comp_si_sl_config_app}
To quantitatively assess the performance difference between the SL and SI algorithms, we conducted paired t-tests on their mean survival rates. This comparison was made at specific trial points and averaged across the three learning phases (Phase 1: trials 1–20, Phase 2: trials 21–60, and Phase 3: trials 61–120).

\begin{table}[hbt!]
\centering
\caption{Paired t-test results comparing mean survival of SL and SI across 15 randomized grid configurations (N=15, df=14). Each condition included 200 agents. Trials 20, 40, and 110 correspond to early, mid, and late points in the experiment, respectively. Phase 1 spans Trials 1–20, Phase 2 spans Trials 21–60, and Phase 3 spans Trials 61–120. Cohen's \textit{d} quantifies effect size, where 0.2, 0.5, and 0.8 are conventionally interpreted as small, medium, and large effects, respectively.}
\label{tab:robustness_ttest_sl_si}
\begin{tabular}{lccccccc}
\toprule
\textbf{Comparison} & \textbf{SL Mean} & \textbf{SI Mean} & \textbf{Diff.} & \textbf{95\% CI} & \textbf{t(14)} & \textbf{p-value} & \textbf{Cohen's d} \\
\midrule
Trial 20          & 23.71 & 23.08 & +0.63 & [0.26, 1.01] & 3.63 & \textbf{0.003}  & 0.94 \\
Trial 40          & 26.23 & 25.04 & +1.19 & [0.23, 2.16] & 2.65 & \textbf{0.019}  & 0.68 \\
Trial 110         & 29.25 & 29.35 & -0.10 & [-0.68, 0.48] & -0.36 & 0.725  & -0.09 \\
\midrule
Phase 1           & 23.02 & 22.75 & +0.27 & [0.16, 0.38] & 5.39 & \textbf{$<$0.001} & 1.39 \\
Phase 2           & 25.91 & 25.09 & +0.82 & [0.37, 1.28] & 3.90 & \textbf{0.002}  & 1.01 \\
Phase 3           & 28.91 & 28.67 & +0.24 & [-0.01, 0.48] & 2.09 & 0.055  & 0.54 \\
\bottomrule
\end{tabular}
\end{table}

\begin{figure}[htbp]
    \centering

    % ---------------- Row 1 ----------------
    \begin{subfigure}[b]{0.31\textwidth}
        \centering
        \includegraphics[width=\textwidth]{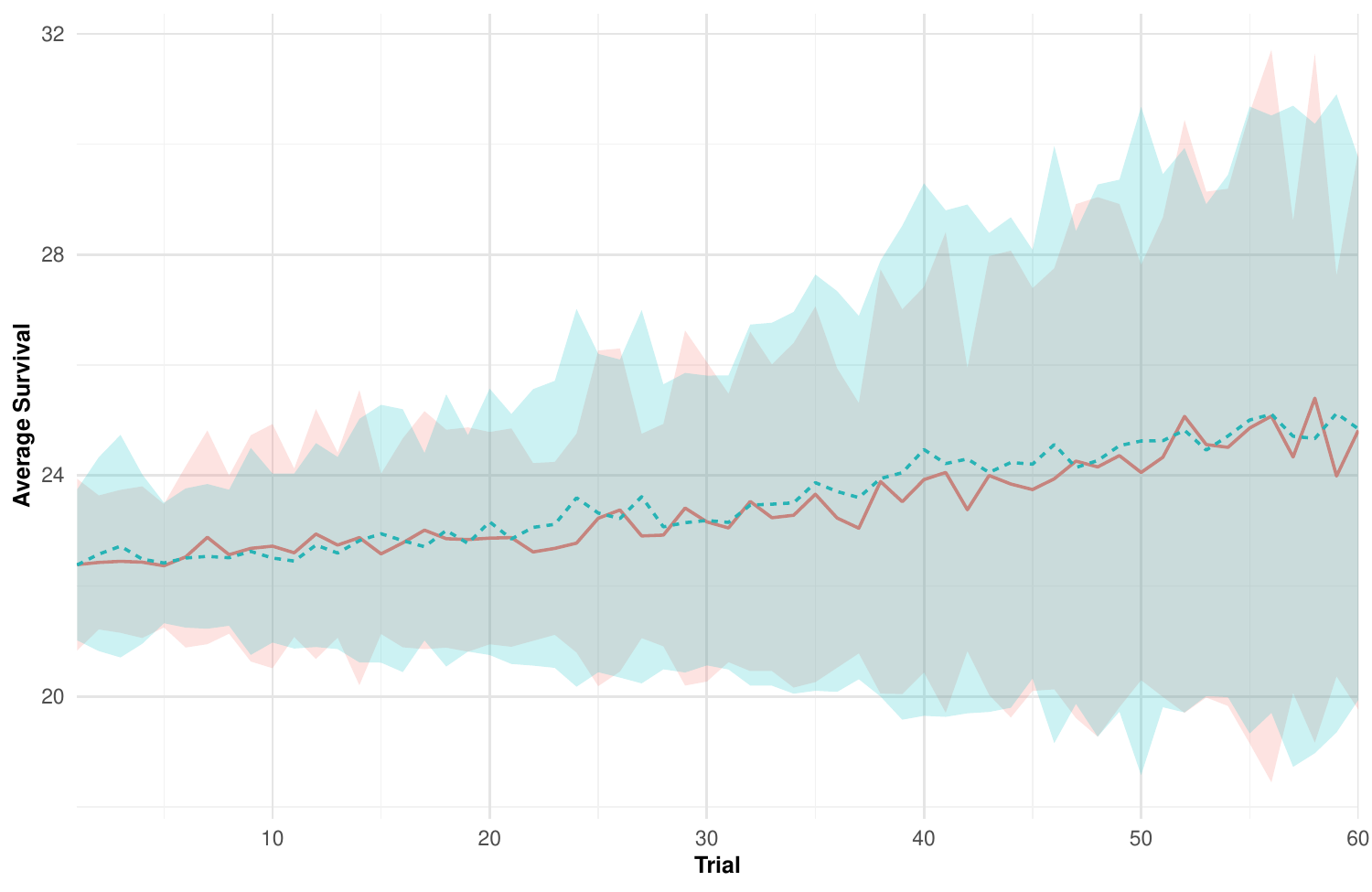}
        \caption{Short-ID 1}
        \label{fig:grid_01}
    \end{subfigure}
    \hfill
    \begin{subfigure}[b]{0.31\textwidth}
        \centering
        \includegraphics[width=\textwidth]{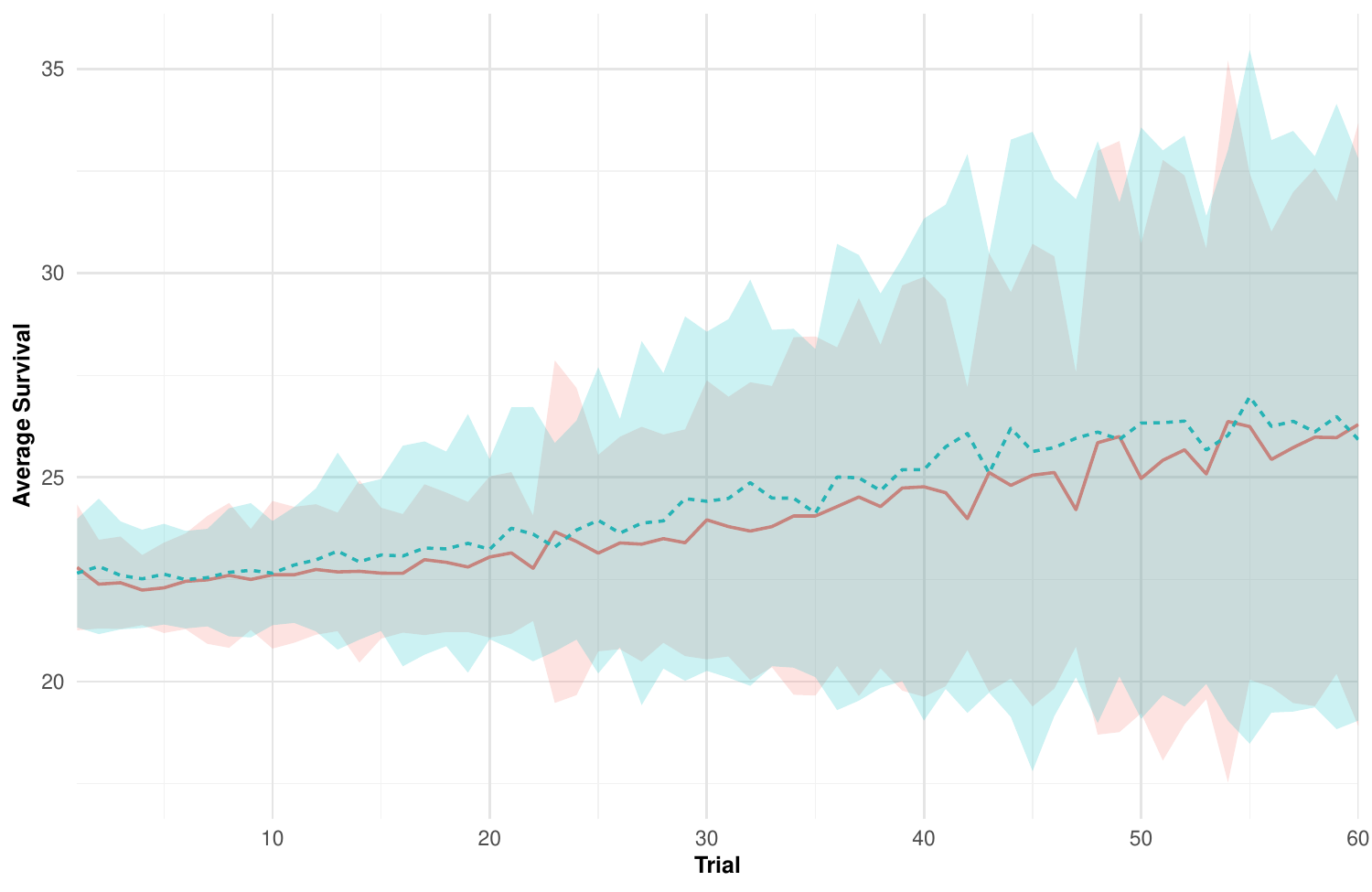}
        \caption{Short-ID 2}
        \label{fig:grid_02}
    \end{subfigure}
    \hfill
    \begin{subfigure}[b]{0.31\textwidth}
        \centering
        \includegraphics[width=\textwidth]{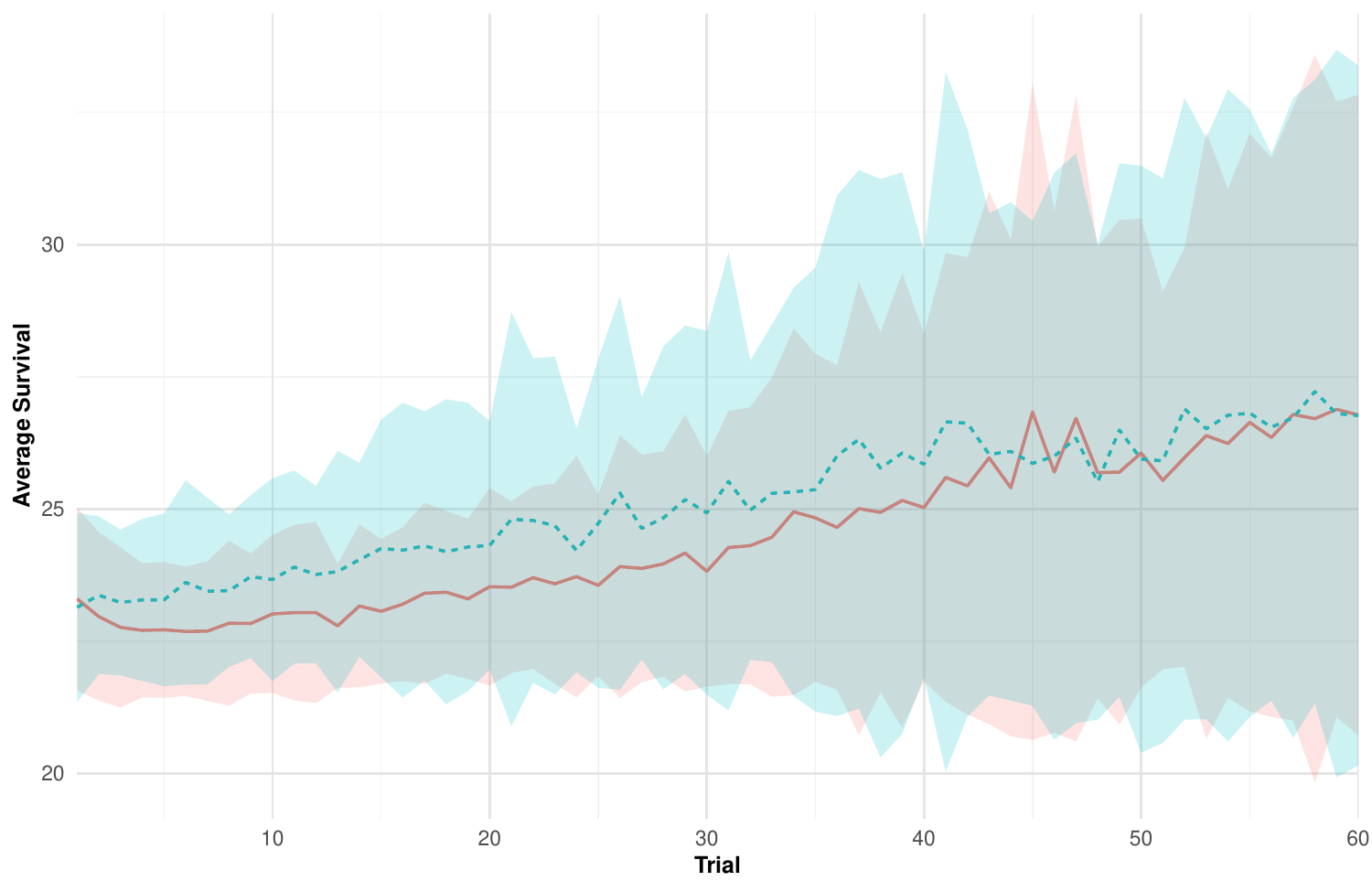}
        \caption{Short-ID 3}
        \label{fig:grid_03}
    \end{subfigure}

    \vspace{1em}

    % ---------------- Row 2 ----------------
    \begin{subfigure}[b]{0.31\textwidth}
        \centering
        \includegraphics[width=\textwidth]{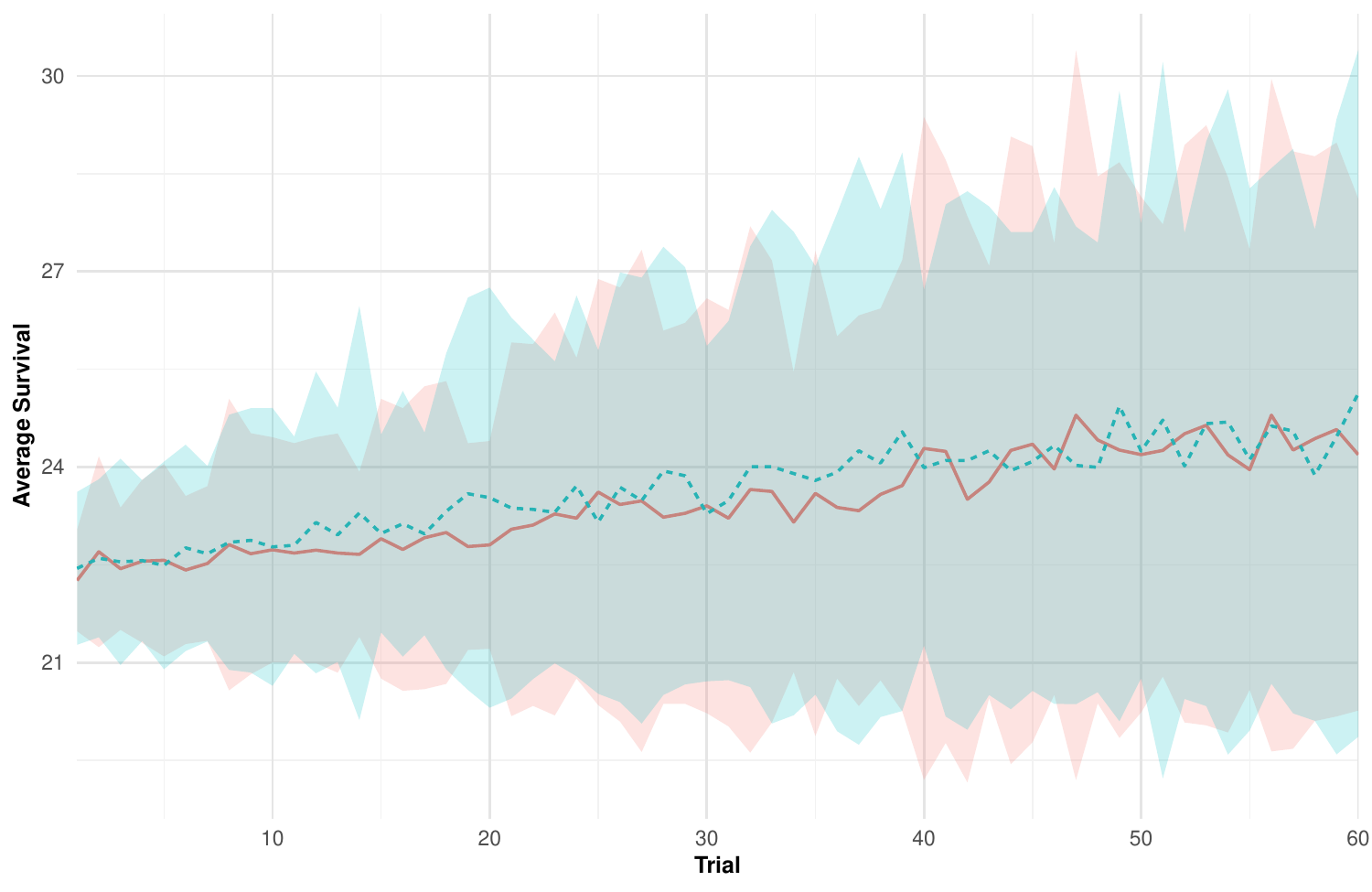}
        \caption{Short-ID 4}
        \label{fig:grid_04}
    \end{subfigure}
    \hfill
    \begin{subfigure}[b]{0.31\textwidth}
        \centering
        \includegraphics[width=\textwidth]{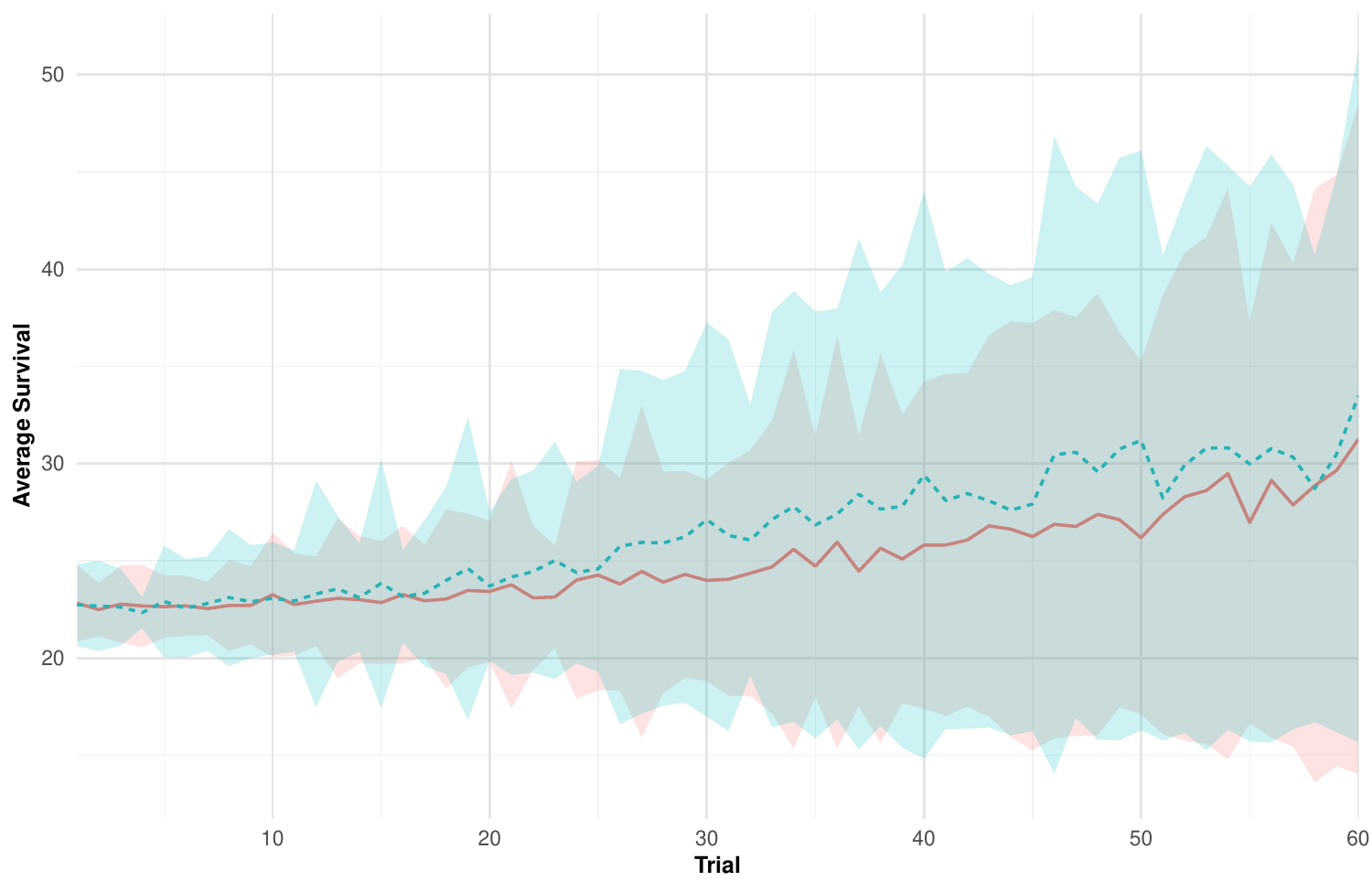}
        \caption{Short-ID 5}
        \label{fig:grid_05}
    \end{subfigure}
    \hfill
    \begin{subfigure}[b]{0.31\textwidth}
        \centering
        \includegraphics[width=\textwidth]{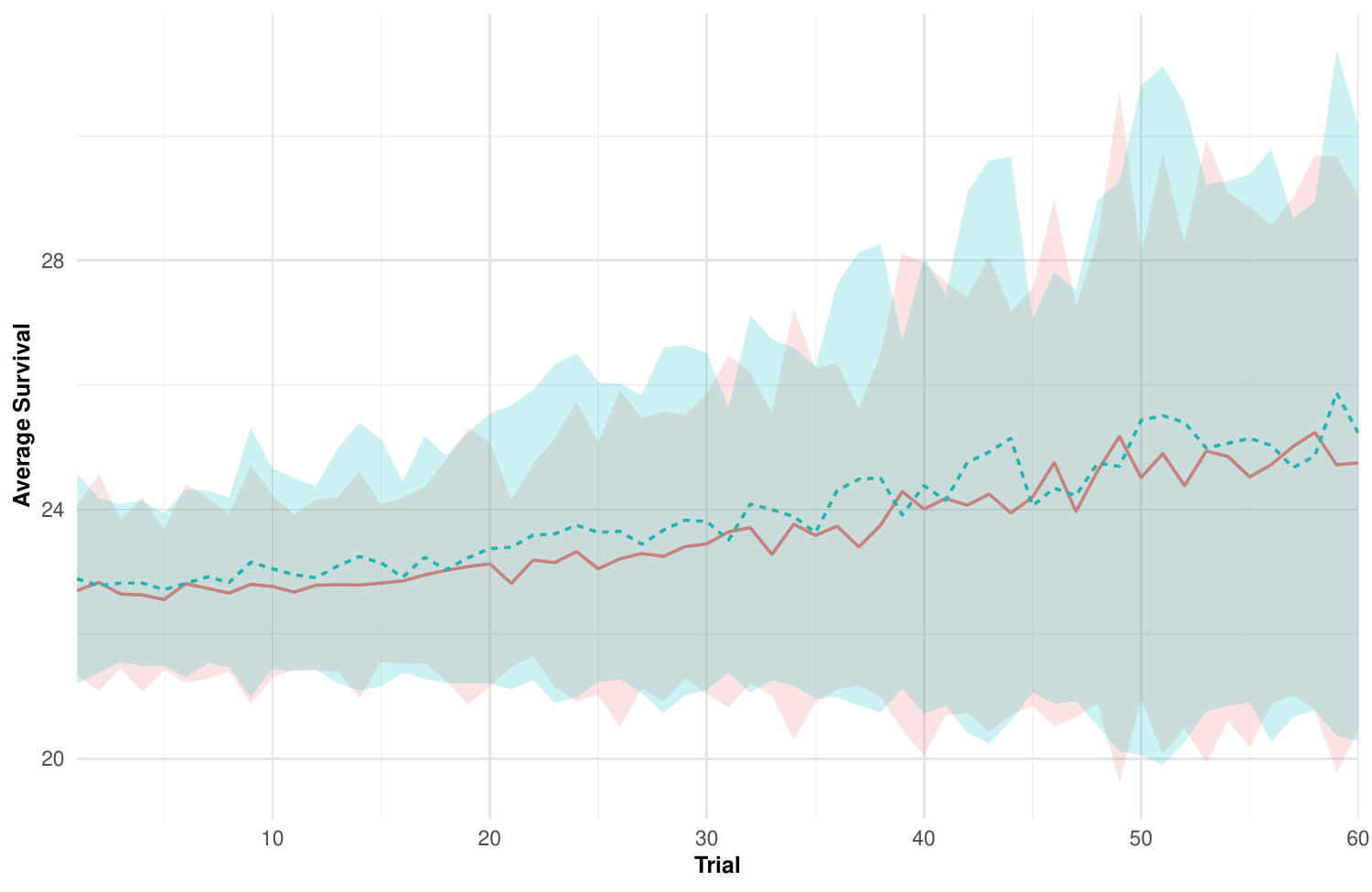}
        \caption{Short-ID 6}
        \label{fig:grid_06}
    \end{subfigure}

    \vspace{1em}

    % ---------------- Row 3 ----------------
    \begin{subfigure}[b]{0.31\textwidth}
        \centering
        \includegraphics[width=\textwidth]{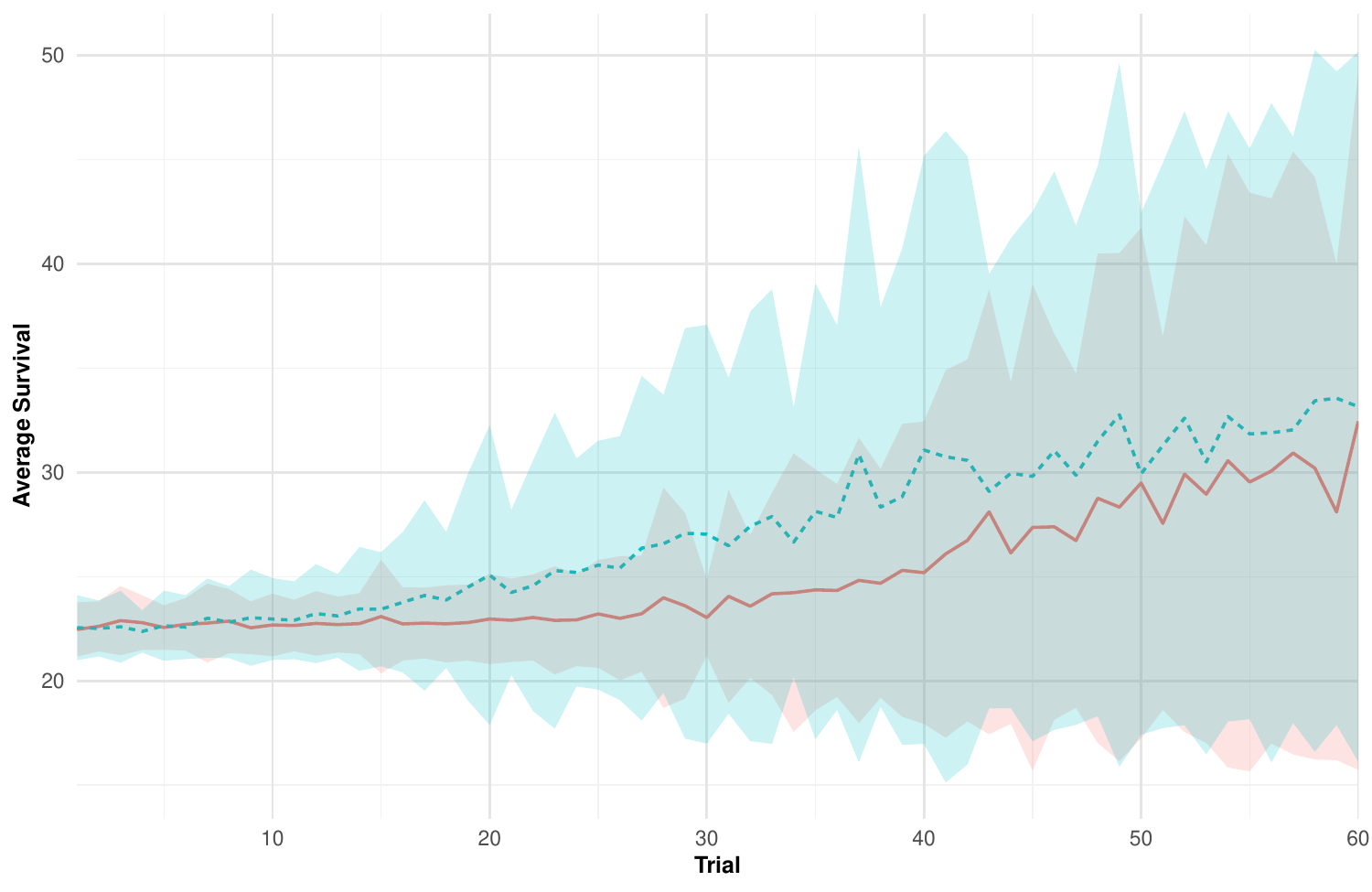}
        \caption{Short-ID 7}
        \label{fig:grid_07}
    \end{subfigure}
    \hfill
    \begin{subfigure}[b]{0.31\textwidth}
        \centering
        \includegraphics[width=\textwidth]{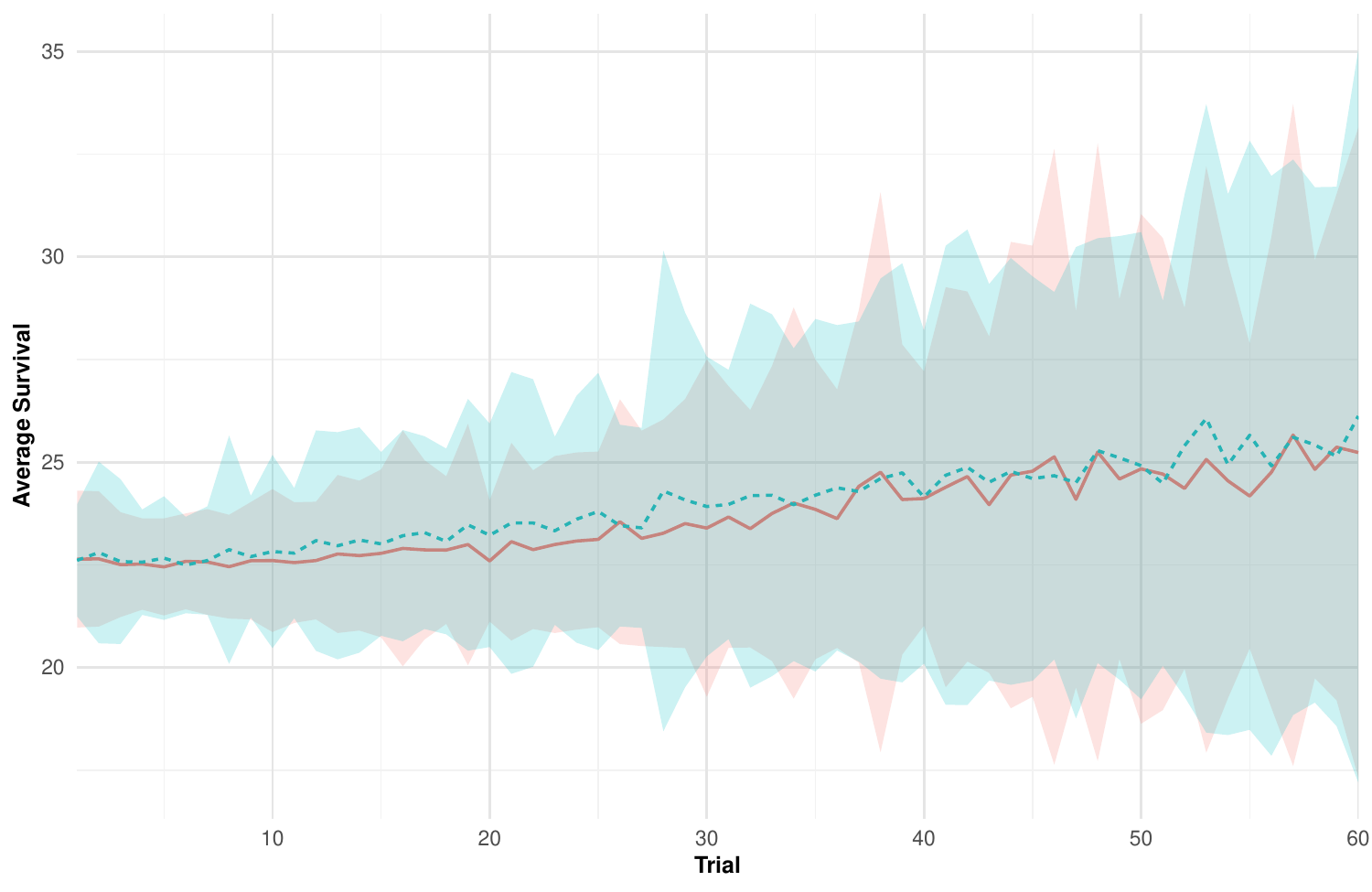}
        \caption{Short-ID 8}
        \label{fig:grid_08}
    \end{subfigure}
    \hfill
    \begin{subfigure}[b]{0.31\textwidth}
        \centering
        \includegraphics[width=\textwidth]{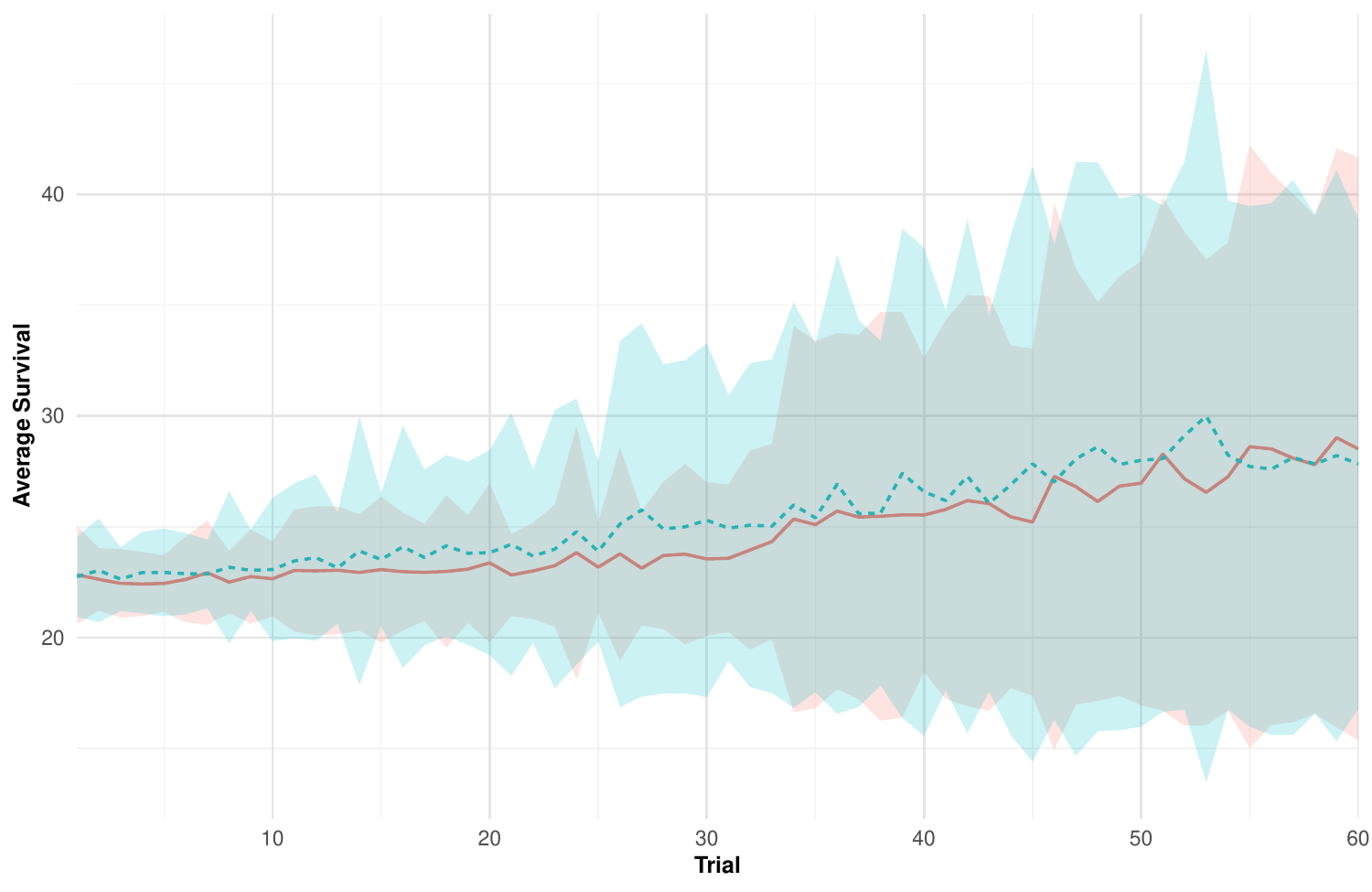}
        \caption{Short-ID 9}
        \label{fig:grid_09}
    \end{subfigure}

    \vspace{1em}

    % ---------------- Row 4 ----------------
    \begin{subfigure}[b]{0.31\textwidth}
        \centering
        \includegraphics[width=\textwidth]{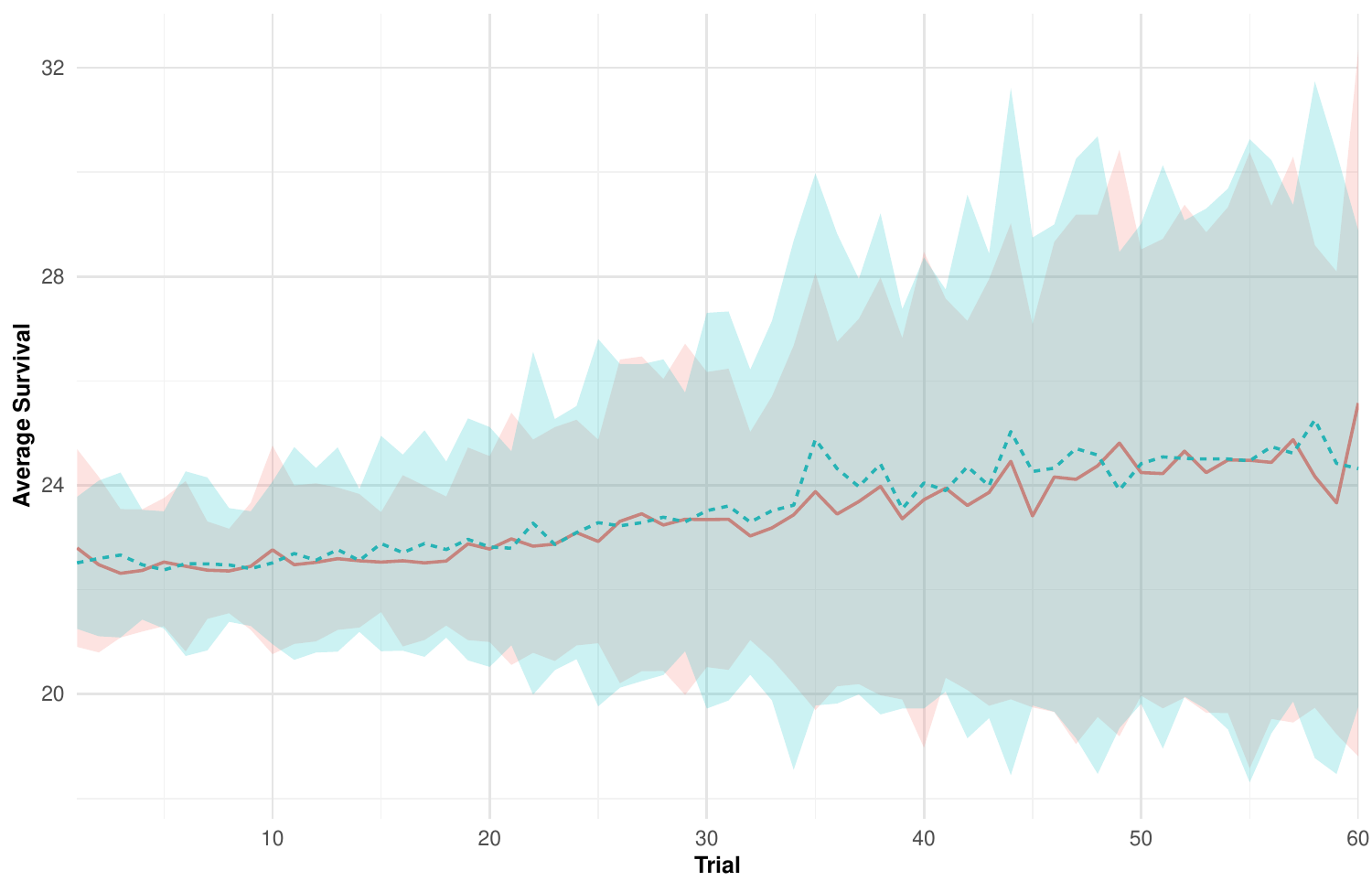}
        \caption{Short-ID 10}
        \label{fig:grid_10}
    \end{subfigure}
    \hfill
    \begin{subfigure}[b]{0.31\textwidth}
        \centering
        \includegraphics[width=\textwidth]{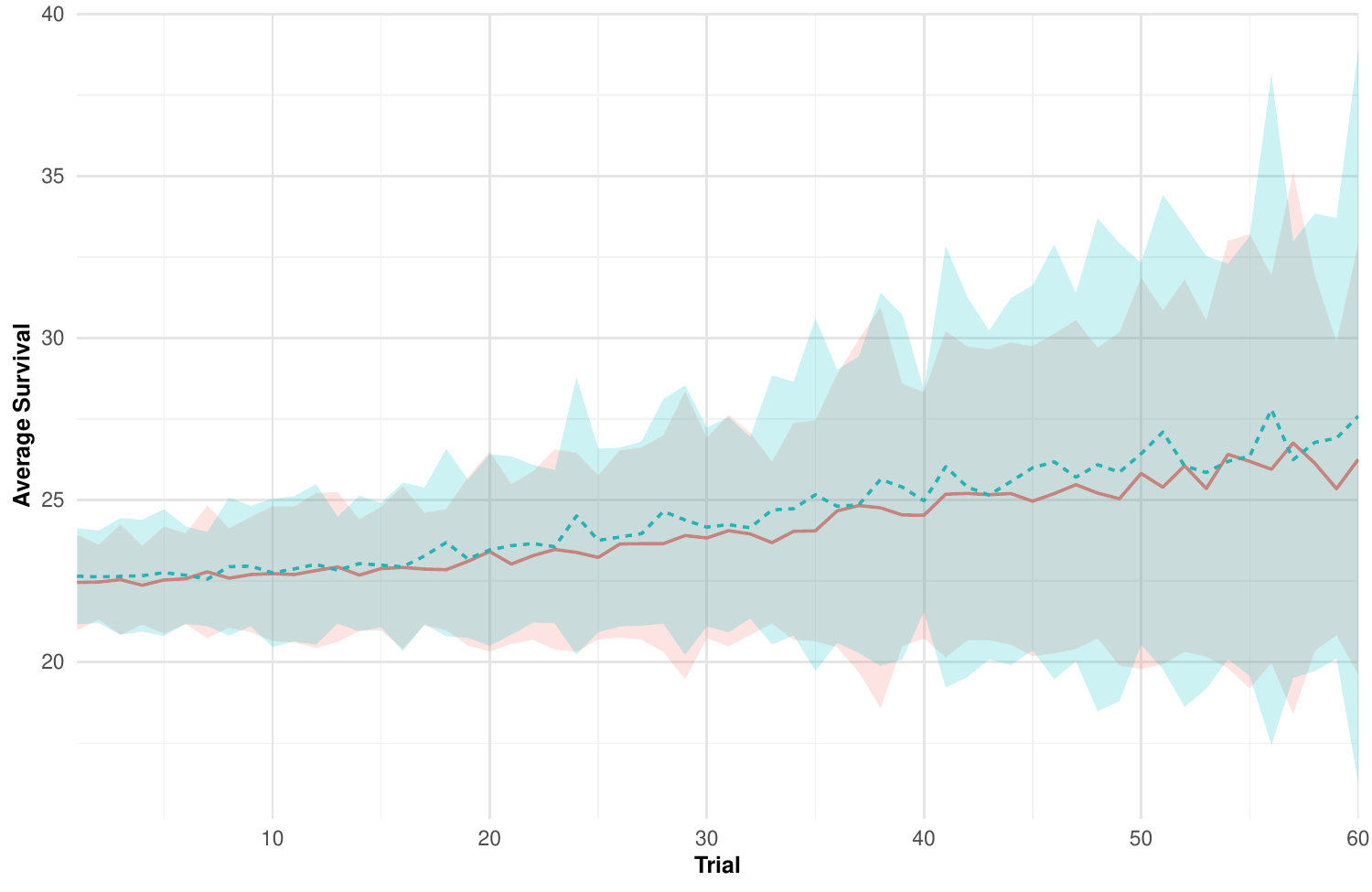}
        \caption{Short-ID 11}
        \label{fig:grid_11}
    \end{subfigure}
    \hfill
    \begin{subfigure}[b]{0.31\textwidth}
        \centering
        \includegraphics[width=\textwidth]{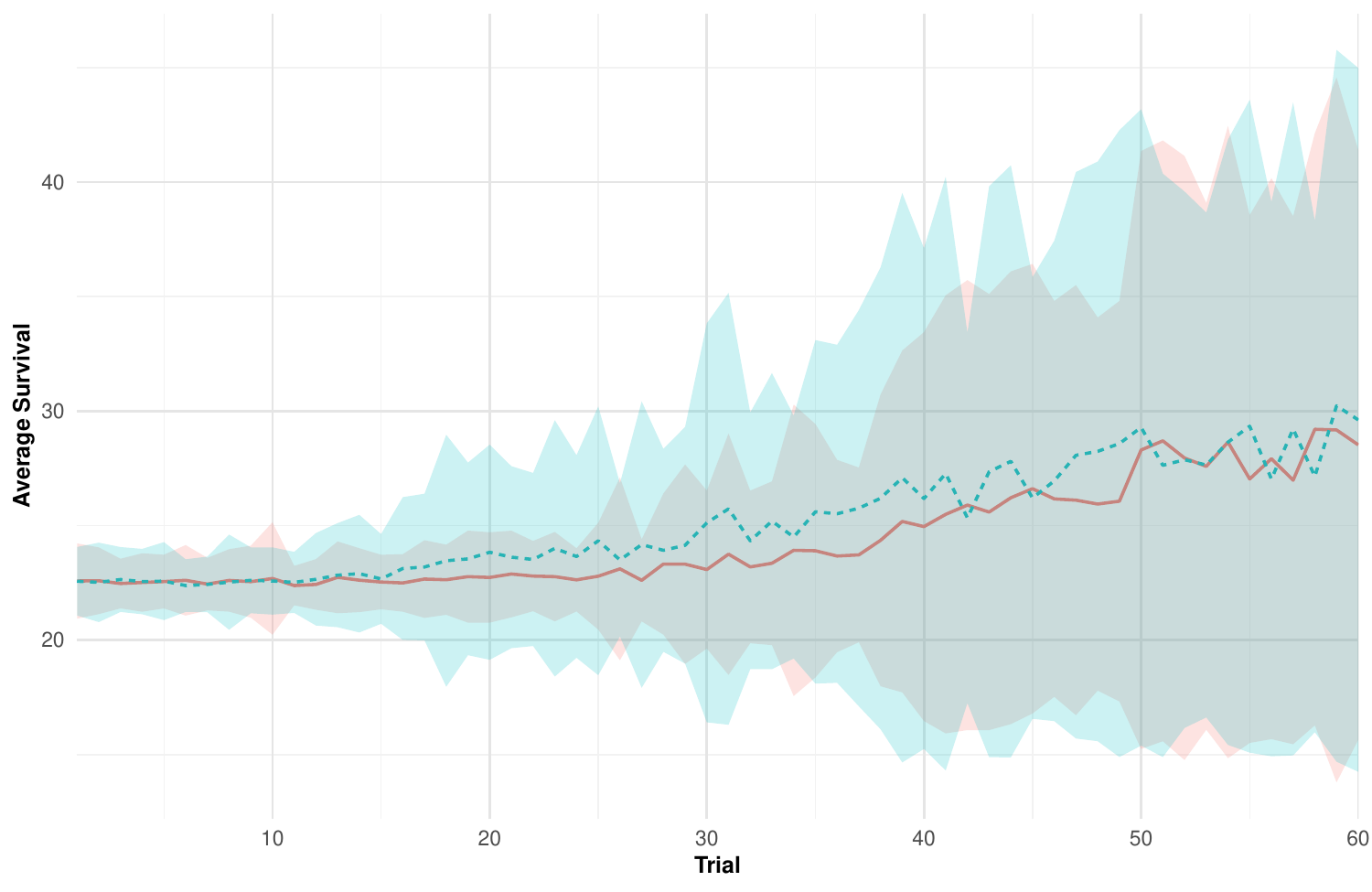}
        \caption{Short-ID 12}
        \label{fig:grid_12}
    \end{subfigure}

    \vspace{1em}

    % ---------------- Row 5 ----------------
    \begin{subfigure}[b]{0.31\textwidth}
        \centering
        \includegraphics[width=\textwidth]{figures/figures_stjohn/grid_robustness_60trials/Plot_Selected_GridID_12d8ff5092adc5fbbc70643df33e88d4_Survival_1to60trials.pdf}
        \caption{Short-ID 13}
        \label{fig:grid_13}
    \end{subfigure}
    \hfill
    \begin{subfigure}[b]{0.31\textwidth}
        \centering
        \includegraphics[width=\textwidth]{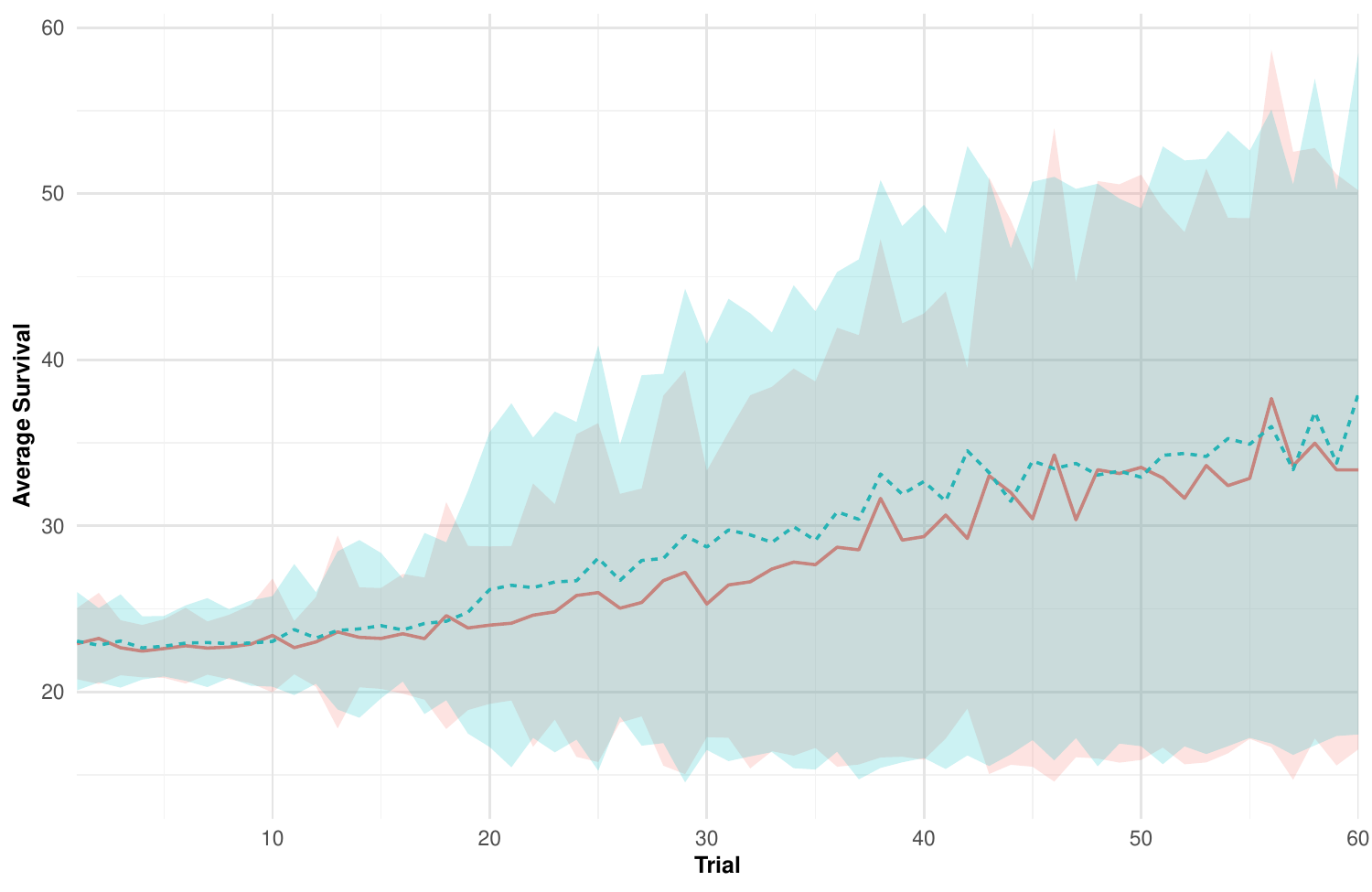}
        \caption{Short-ID 14}
        \label{fig:grid_14}
    \end{subfigure}
    \hfill
    \begin{subfigure}[b]{0.31\textwidth}
        \centering
        \includegraphics[width=\textwidth]{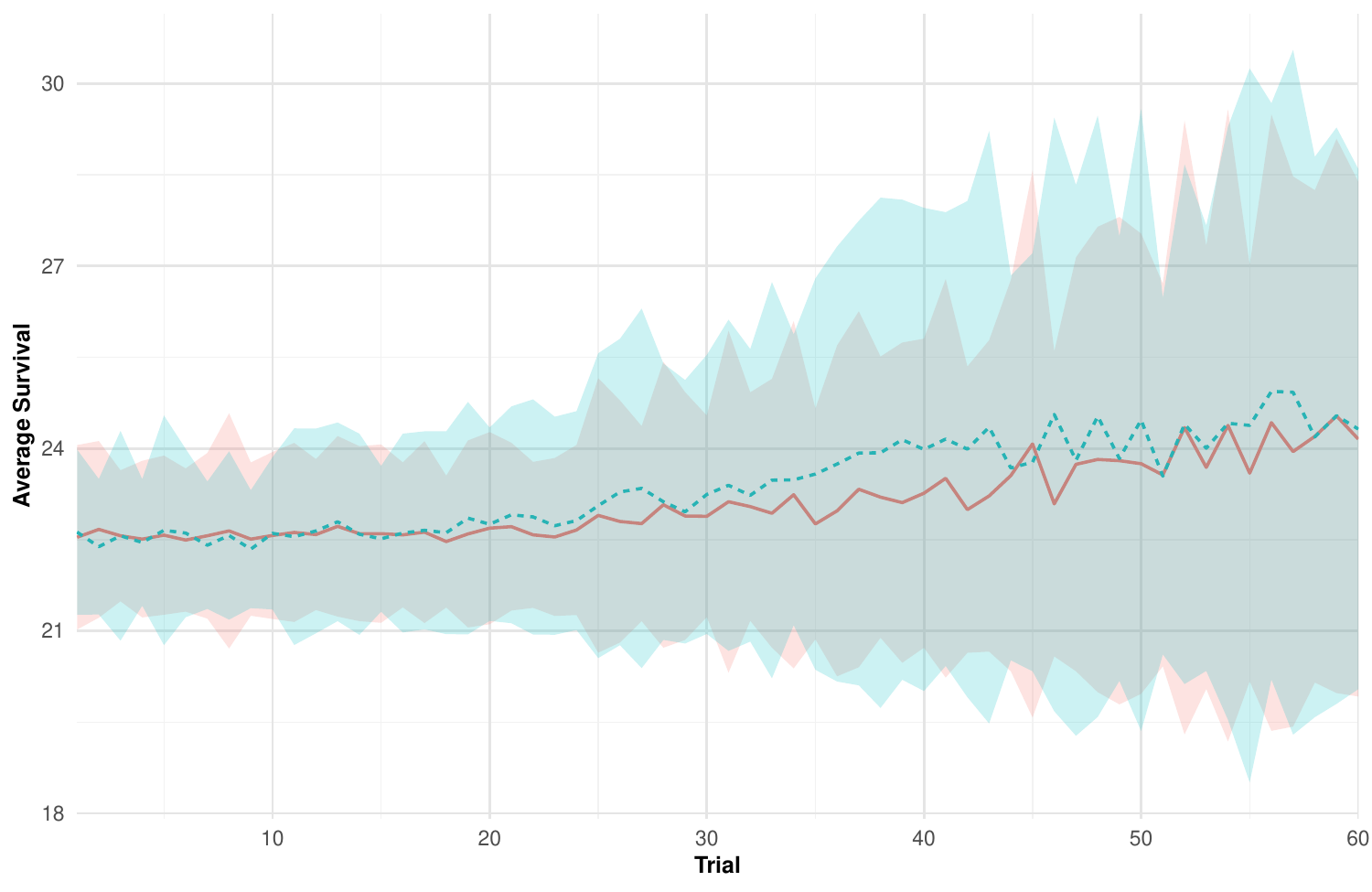}
        \caption{Short-ID 15}
        \label{fig:grid_15}
    \end{subfigure}

    \caption{Early-phase survival curves for fifteen randomly sampled grid environments used in analyses of the robustness (generalizability) of observed performance differences between algorithms. Each plot shows the average survival across 60 trials for the SI and SL algorithms, aggregated over 200 random seeds per algorithm. Variability is visualized using standard deviation bands. Although the error bands could be further tightened with more runs, the observed early learning advantage — when interpreted alongside the statistical comparisons presented in the main text — provides robust evidence for the effect.}
    \label{fig:combined_grid_robustness_15}
\end{figure}

As shown in Table \ref{tab:robustness_ttest_sl_si}, results from the 15 grid configurations demonstrate a clear and robust advantage for the SL algorithm, particularly in the initial and middle stages of learning. This overall effect is an average across configurations of varying difficulty; in the most challenging layouts where learning was arduous for both agents, the performance difference was naturally muted (e.g. Figure \ref{fig:grid_11}). However, the data clearly show that where learning was more achievable, the advantages shown for SL became pronounced (e.g. Figure \ref{fig:grid_05}). This is reflected in the aggregate statistics for Phase 1 (trials 1–20), where performance in SL was significantly higher than SI — a finding that was both statistically significant (p<0.001) and reflected a large effect size (Cohen’s d=1.39). This early advantage was sustained through Phase 2 (trials 21–60), which also showed a significant difference (p=0.002) with a large effect size (d=1.01).

In later stages of training (Phase 3, trials 61–120), performance differences between the two algorithms narrowed considerably. The difference was no longer statistically significant at the conventional $\alpha = 0.05$ level, but it remained noteworthy ($p = 0.055$). This indicated that while SI eventually catches up, SL may maintain a slight edge on average. These findings strongly suggest that the structured, active learning in SL confers a tangible and generalizable advantage in early-phase adaptation across unfamiliar environments. This is likely due to its intrinsic drive to revisit informative states (e.g., the hill) to rapidly build and refine its model. While both algorithms converge toward similar high-performance levels over extended training, SL consistently exhibited a faster and more efficient adaptation profile, consistent with our main results.

These results are also in line with a broader theoretical argument that planning depth should interact with spatial structure. For example, in a $10 \times 10$ grid, the expected Manhattan distance between random points is about 6.6 steps:
\[
E[d_{\text{Manhattan}}] = 2 \cdot \frac{N^2 - 1}{3N} \approx 6.6 \quad \text{when} \quad N=10.
\]
As discussed in Section~\ref{horiz_analysis_res}, this limits the gains from planning beyond 4–6 steps, and algorithms that adapt within this horizon tend to perform best. SL addresses this constraint effectively; in contrast, SI and BARL lack explicit mechanisms to prioritize strategic information gathering and may therefore lag in early adaptation.

\subsection{Mechanisms of Differential Performance}
\label{subsubsec:differential-performance}
\begin{figure}[h]
\centering
\includegraphics[width=1\textwidth]{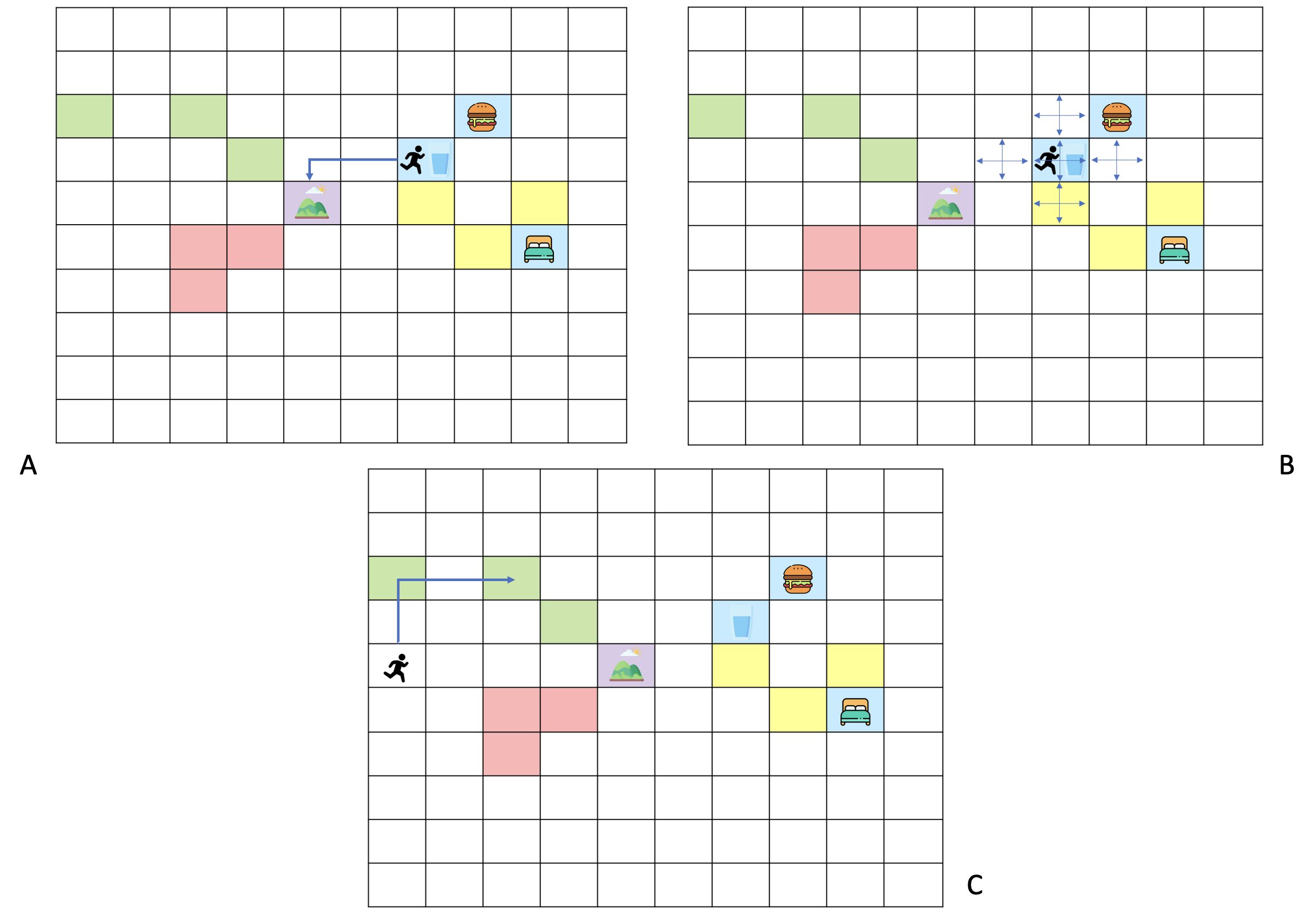}
\caption{\label{fig:example1} (\textbf{A}) The SL agent frequently visited the hill within a trial to contextualize the information it learned with greater accuracy (e.g., observing water and then quickly moving to the hill to figure out what the associated context was). (\textbf{B}) The SI agent did not see value in the hill with respect to parameter learning, and so did not frequently revisit this state to contextualize its prior observations. It instead continued to explore the state-space more broadly to seek out novel observations. (\textbf{C}) \textit{Supplementary demonstration of parameter dependence}: Here we note that, if the precision of the preference distribution was sufficiently high, both SI and SL agents often ignored the hill and immediately attempted to guess where resources might be. This was because very high preference precision values effectively down-weight the other terms in the EFE and thereby deter exploration (see main text for details).}
\end{figure}

Further investigation clarified the mechanisms underlying differential performance between the four algorithms. Note that, in order to accurately learn the likelihood mapping, it is crucial that the agent discover resource positions through random exploration, and then subsequently associate them with the correct context. In this case, the SL agent showed faster improvement over time because, unlike the other three algorithms, it would often immediately prioritize moving to the hill after discovering a resource. This allowed it to better connect the context (season) with the location of the resource (shown in Figure \ref{fig:example1} \textbf{A}). SI learned at a somewhat slower rate, as it did not recognize the hill as a valuable state to visit after discovering a resource. While SI did value the hill state in these simulations, it only did so when either its beliefs about context became sufficiently imprecise or when the model was precise and the hill became a Bellman-optimal option. Thus, it would not necessarily do so after unexpectedly discovering a resource. 

Figure \ref{fig:example1} \textbf{B} shows example behavior of an SI agent when finding a resource. Unlike SL, this agent does not prioritize moving to the hill. Instead, it continues to explore surrounding states that it has not yet visited. In general, it also visits the hill far less frequently. Learning in this way is relatively slower, as the SI agent, on average, has less precise beliefs about the context in which it has observed a resource. Both BARL algorithms showed very poor learning, as they did not value the hill when their models were imprecise. This is because they did not view the hill as providing a Bellman-optimal mechanism for moving to a resource (as they might if the environment was fully known). The logic here is that, if the agent had imprecise beliefs over which context mapped to which resource position, there was no point in learning what the context was before attempting to move to a resource. In the case of BARL without UCB, learning was primarily chance-based (i.e., when the agent randomly visited a resource location or visited the hill on the way to some other location it viewed as rewarding). The addition of UCB was also unhelpful. To see why, it is first useful to consider that, while UCB provided a directed exploration drive, this was not \textit{goal-directed} in the manner displayed by SL. That is, while UCB-driven exploration was directed at states for which the agent had made few observation (i.e., and therefore not random), it was not driven like SL by the goal of precise `credit assignment' of resource locations to a season (i.e., its exploration heuristic did not encourage it to move to the hill to resolve context ambiguity upon discovering a resource). Rather, its exploration was entirely based on relative frequency of states visited over time. In the present environment, which had fairly sparse rewards, this resulted in somewhat wasteful over-exploration of states without resources or epistemic value (regarding the current season).

\subsubsection{Model Learning Analysis for SL}

\begin{figure}[h]
    \centering
    \includegraphics[width=0.85\linewidth]{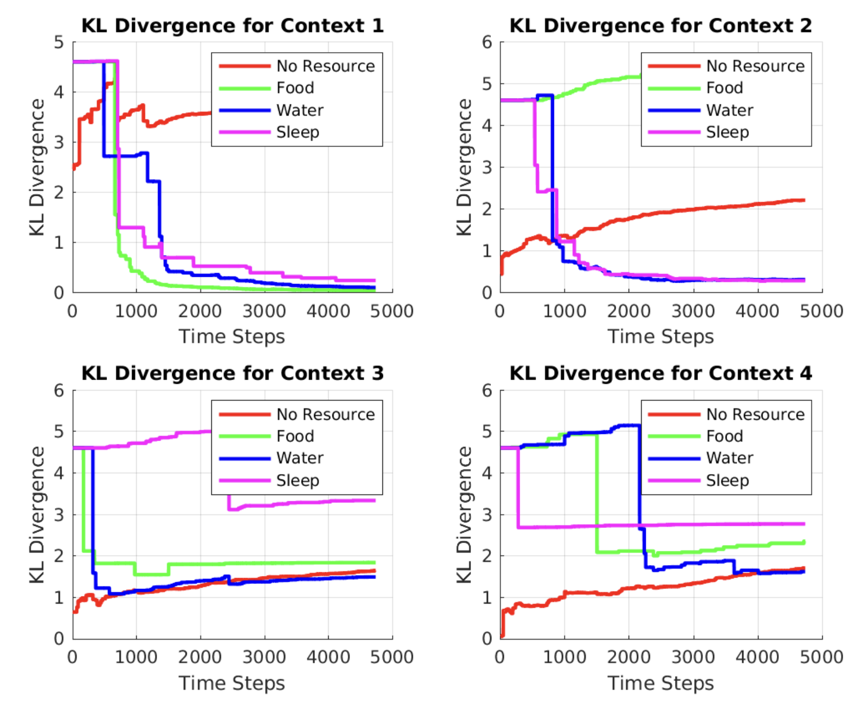}
    \caption{\textbf{KL divergence between the SL agent’s posterior beliefs and the true resource distributions, plotted over cumulative time steps aggregated across all trials.} Each subplot corresponds to one of the four contexts (seasons), with separate curves for each resource modality (Food, Water, Sleep) and the 'No Resource' observation. A declining KL divergence indicates improved learning accuracy. While the agent achieves strong convergence in Season 1, learning remained unstable or incorrect in Seasons 2--4 with respect to some modalities (e.g., Food in Season 2, Sleep in Season  3), suggesting persistent belief miscalibration and context-specific learning failures.}
    \label{fig:divergence_example}
\end{figure}

Figure \ref{fig:divergence_example} shows further analyses to better understand the learning dynamics in SL shown above. These results indicated that, when learning went poorly, it was often context- and modality-specific. For example, in the first season, SL showed accurate learning for all resource locations (based on the KL-divergence between posterior beliefs and ground truth), whereas the second season showed accurate learning for Sleep and Water, but inaccurate learning for Food locations. 

% \subsection{Further Notes}
% \label{app:computational_formalisms}
\subsection{Memoization and Computational Efficiency}
\label{app:memoization}
To mitigate the computational burden of the recursive tree search inherent in SI and SL, we employed memoization. Memoization involves caching and retrieving the EFE (or reward, for BARL) values for previously encountered state-action pairs or sub-trees during the planning process. This technique involves storing previously calculated EFE values of a subset of search trajectories in memory, allowing reuse when future searches overlap with already-calculated subsets.

While memoization significantly reduces the number of redundant computations, especially for deeper searches in environments with overlapping state trajectories (see horizon analysis, Figure~\ref{fig:memory-tree-search-analysis}), it is not without caveats. For ActInf agents, caching EFE values based on a subset of state factors can introduce inaccuracies if path-dependent components of belief (e.g., the history influencing current uncertainty about model parameters, which in turn affects the epistemic/novelty value) are not fully captured in the cache key. Furthermore, a notable drawback to memoization is the significant memory requirement to store these cached values. However, if needed for future applications, this memory burden could be mitigated via various approximation techniques, such as Coarse Coding \citep{Sutton2018}, a form of linear function approximation, or Artificial Neural Networks \citep{Oudare}. The choice to use memoization represents a trade-off between computational efficiency, memory usage, and the precision of the EFE approximation. Our use of memoization was a pragmatic choice for managing the extensive simulations required. For the interested reader, below we show the number of recursive function calls made with vs. without memoization from $t = 1$. Results under different possible search depths are also included.

\begin{table}[htbp]
\centering
\caption{A comparison of the number of recursive function calls with and without memoization.}
\begin{tabular}{lrr}
\toprule
\textbf{Search Depth}& \textbf{With Memoization} & \textbf{Without Memoization} \\ 
\midrule
0 & 1   & 1     \\
1 & 17  & 21    \\
2 & 53  & 421   \\
3 & 117 & 8421  \\
4 & 257 & 160421 \\
\bottomrule
\end{tabular}
\end{table}

\end{document}